\documentclass[make,article,accept,pdftex,moreauthors]{Definitions/mdpi} 
\firstpage{1}
\pubvolume{1}
\issuenum{1}
\articlenumber{0}
\pubyear{2026}
\copyrightyear{2026}
\externaleditor{~} 
\datereceived{26 May 2026}
\daterevised{26 June 2026} 
\dateaccepted{29 June 2026}
\datepublished{ }

\usepackage{mathtools}
\usepackage[shortcuts]{extdash}

\DeclareMathOperator*{\argmin}{arg\,min}

\providecommand{\Description}[1]{}
\externalbibliography{yes}

\usepackage{xcolor}

\Title{Deformable Medical Image Registration with KAN-Based Implicit Neural Representations}

\Author{Nikita 
 A. Drozdov$^{1,2}$, Marat O. Zinovev$^{1,2}$ and Dmitry V. Sorokin$^{1,2}$ *}

\AuthorNames{Nikita A. Drozdov, Marat O. Zinovev and Dmitry V. Sorokin}

\address {%
$^{1}$ \quad AI Center, Lomonosov Moscow State University, 
 119991, Moscow, 
 Russia\\
$^{2}$ \quad Laboratory of Mathematical Methods of Image Processing, Faculty of Computational Mathematics and Cybernetics, Lomonosov Moscow State University,
 119991, Moscow,
 Russia; drozdovna@my.msu.ru (N.A.D.); zinovevmo@my.msu.ru~(M.O.Z.)}

\corres{\hangafter=1 \hangindent=1.05em \hspace{-0.82em}Correspondence: dsorokin@cs.msu.ru~(D.V.S.)}

\abstract{Deformable image registration (DIR) is central to medical image analysis, supporting spatial alignment for longitudinal studies and multi-modal fusion. Learning-based methods such as CNNs and transformers provide rapid inference but often require large training datasets and can underperform classical iterative methods for specific anatomies or modalities. Implicit neural representations (INRs) offer a data-efficient alternative by modeling deformation fields as continuous coordinate-to-displacement mappings, yet their per-pair optimization makes runtime efficiency and robustness to initialization essential. We introduce KAN-IDIR and RandKAN-IDIR, the first Kolmogorov--Arnold network (KAN)-based INR framework for pairwise-optimized, resolution-independent DIR, designed to improve seed stability and resource efficiency without requiring a large training dataset. KANs use learnable activation functions that are well suited to continuous, physically structured deformation fields. RandKAN-IDIR further reduces cost through randomized basis sampling, preserving registration quality with fewer basis functions. We evaluate the methods on lung CT, brain MRI, and cardiac MRI datasets against pairwise-optimized neural approaches, dataset-trained deep models, and classical baselines. KAN-IDIR and RandKAN-IDIR provide the strongest overall performance among pairwise-optimized neural registration methods across all three datasets, with low computational overhead and superior stability across random initializations. On ACDC, KAN-IDIR also achieves the highest DSC and best deformation regularity among all compared methods. RandKAN-IDIR slightly outperforms adaptive basis selection variants while avoiding their additional training-time complexity. This makes the approach practical for reproducible clinical research use. Source code is publicly available.} 


\begin{document}

\section{Introduction}

Deformable image registration (DIR) is a fundamental task in the field of medical image processing. Its primary goal is to determine a spatial transformation that aligns one image with another, ensuring that corresponding regions appear in the same locations. This process enables accurate comparison, fusion, or further analysis of imaging data.

Before the advent of deep learning, non-rigid registration relied on iterative optimization techniques. Examples of methods following this approach include Free-Form Deformation~\citep{Rueckert1999}, LDDMM~\citep{Beg2005LDDMM}, Demons~\citep{Thirion1998}, Elastix~\citep{Klein2009Elastix}, ANT~\citep{Avants2011ANTs}, pTV~\citep{vishnevskiy2016isotropic}, and Flash~\citep{Zhang2019FourierFLASH}. These approaches are well-established and mathematically grounded, but they require tuning parameters for each image pair and involve substantial computational costs, which makes them less suitable for real-time registration of large images.
With recent advances in artificial intelligence technologies, image-to-image architectures such as convolutional neural networks (CNNs) and transformers have become increasingly popular in medical image registration \citep{balakrishnan2019tmi, chen2022transmorph}. These approaches predict displacement vector fields for image pairs, offering inference speeds several orders of magnitude faster than classical iterative methods---an essential advantage for real-time clinical applications.
However, such learning-based methods face three primary limitations: (1) their performance relies on large training datasets, which are often unavailable for specific organs or imaging modalities; (2)~they do not consistently outperform classical methods for certain anatomical structures; and (3) they exhibit a degradation of registration quality when input data resolution varies.
An alternative approach to dataset-trained registration methods is to learn the deformation field as a mapping from continuous voxel coordinates to the corresponding displacement vector. This mapping is parametrized by a neural network, referred to as an implicit neural representation (INR). INRs have demonstrated remarkable performance in tasks like novel view synthesis (NeRFs) \citep{mildenhall2020nerf} and have been leveraged in several methods for DIR \citep{wolterink2022implicit, vanharten2023robust, vanharten2024deformable}. 

INR-based solutions address key limitations of dataset-trained deep learning models: they reduce dependence on large training sets, are less sensitive to domain shifts, and naturally handle resolution changes. However, because they optimize per image pair, their scientific and clinical utility depends on amount of compute and memory, seed-stable optimization (to support reproducible studies on small cohorts), and deformation regularity (to avoid confounding downstream measurements).
Most existing INR architectures rely on multi-layer perceptrons (MLPs) with fixed activation functions \citep{sitzmann2020siren, saragadam2023wire, Liu_2024_CVPR}. Recently, Kolmogorov--Arnold Networks (KANs), with learnable activations, have emerged as a promising alternative to traditional MLPs in regression tasks \citep{liu2024kan}. Several recent studies have examined KAN-based INRs across various tasks \citep{mehrabian2024implicit, li2025representing}. 
To date, however, KAN-based INRs for deformable medical image registration, especially in multi-organ benchmarks with pairwise optimization, remain largely unexplored.

In this paper, we propose a novel approach to DIR using Implicit Neural Representations enhanced with Kolmogorov--Arnold Networks. 
Our key contributions are:

\begin{enumerate}
\item We 
 introduce the first KAN-based INR framework targeting pairwise-optimized, resolution-independent DIR with an explicit focus on seed stability and resource efficiency, instantiated as KAN-IDIR and RandKAN-IDIR with randomized basis sampling. Our method achieves accurate deformations while keeping computational and memory costs low compared to classical and neural network-based baselines.

\item Among pairwise-optimized neural registration methods in our comparison, KAN-IDIR and RandKAN-IDIR provide the strongest overall performance across all three datasets, spanning different organs and imaging modalities. On DIR-Lab (lung CT) and OASIS-1 (brain MRI), dataset-specific or dataset-trained methods such as pTV and TransMorph remain stronger overall baselines, whereas on ACDC (cardiac MRI) KAN-IDIR achieves the highest DSC and best deformation regularity among all compared methods. In addition, the proposed methods produce fewer large-error outliers compared to existing INR-based baselines on the DIR-Lab dataset.

\item We show that RandKAN-IDIR, which employs randomized basis sampling, achieves performance comparable to or slightly better than an adaptive KAN variant that learns basis function indices, while eliminating the additional training-time computational overhead associated with adaptive basis selection.
\end{enumerate}

\subsection*{Related Work}

\textbf{Implicit Neural Representations. 
} Implicit neural representations model signals as continuous, differentiable functions parameterized by neural networks. Following \citep{sitzmann2020siren}, an INR is defined as the solution to an equation: 

\begin{equation} \label{eq1}
\mathcal{C}(\mathbf{x},\Phi,\nabla_{\mathbf{x}}\Phi,\nabla_{\mathbf{x}}^2\Phi,\ldots) = 0, \quad \Phi:\mathbf{x}\mapsto\Phi(\mathbf{x}).
\end{equation}
Here, 
a network is trained to parameterize the function $\Phi$ that maps coordinates $\mathbf{x} \in \mathbb{R}^n$ to target values while satisfying the implicit constraint $\mathcal{C}$ in Equation (\ref{eq1}).

The choice of INR architecture is critical, as it determines the class of functions that the network can effectively represent. 
While ReLU-based MLPs mainly model low-frequency signals \citep{mildenhall2020nerf}, many works explore alternative activation functions—such as the sine functions in SIREN \citep{sitzmann2020siren}, Gaussian activations \citep{ramasinghe2022beyond}, wavelet-based activations in WIRE \citep{saragadam2023wire}, or variable-periodic functions in FINER \citep{Liu_2024_CVPR}—to better capture high-frequency structure. These designs, however, introduce their own challenges. For example, the expressiveness of SIREN depends heavily on frequency-related hyperparameters and is sensitive to initialization, requiring careful tuning to avoid unstable or inconsistent behavior \citep{vonderfecht2024predicting}. Moreover, instead of relying on hand-crafted periodic or frequency-aware activations, recent work has explored learnable activation functions, whose parameters adapt during training \citep{goyal2019learning}. Spline-based versions, such as those used in KAN \citep{liu2024kan}, offer a flexible way to model nonlinearities and frequency content, helping mitigate some limitations of fixed activation choices.
Nevertheless, SIREN remains a strong and widely used baseline in many domain-specific applications, including medical image registration.

\textbf{Medical Image Registration.} 
Classical iterative image registration methods are built upon well-established and rigorously developed mathematical frameworks. The authors of \citep{Rueckert1999} introduced a non-rigid registration approach using B-spline-based free-form deformations (FFDs) combined with an affine global transformation and normalized mutual information to model complex local tissue motion. LDDMM \citep{Beg2005LDDMM} models large, smooth deformations by interpreting them as geodesic flows, grounded in concepts from fluid mechanics and Lagrangian dynamics. In \citep{Thirion1998}, the authors proposed a model in which one image diffuses through the boundaries of another, analogous to Maxwell’s demons \citep{Maxwell1871}. The pTV method \citep{vishnevskiy2016isotropic} parameterizes the displacement field with B‑splines and incorporates isotropic total variation regularization to better capture breathing motion. Flash \citep{Zhang2019FourierFLASH} accelerates diffeomorphic registration by representing the initial velocity field in a low-dimensional Fourier space, reducing computational cost with minimal loss of accuracy. While these classical approaches achieve high accuracy through per-case optimization, they are often computationally intensive---particularly for high-resolution images---and require extensive hyperparameter tuning to reach optimal performance. These factors limit their scalability and practicality in clinical settings.

Recent advances in deep learning (DL) for medical image registration have been heavily CNN-based methods. The VoxelMorph framework \citep{balakrishnan2019tmi} was a major breakthrough, enabling fast deformable registration using CNNs to learn deformation fields in an unsupervised manner. It was further extended with probabilistic modeling with stationary velocity fields (SVFs) \citep{dalca2019unsupervised}, the diffeomorphic networks in LapIRN approach \citep{mok2020large}, and methods ensuring symmetry and topology preservation \citep{mok2020fast,greer2021icon,tian2023gradicon}. Vision transformer-based methods further advanced DIR by capturing richer spatial dependencies, as demonstrated by TransMorph~\citep{chen2022transmorph} and, more recently, CTCF, which uses a three-level coarse-to-fine Swin Transformer cascade \citep{pasenko2026ctcf}.
Beyond CNNs and transformers, the authors of \citep{meng2024correlation} developed a pure-MLP method, achieving top results on some datasets.

While CNN-based medical image registration has advanced significantly, key limitations persist. First, even when trained on large datasets, these image-to-image models remain vulnerable to domain shifts \citep{jena2024deep}---a problem often addressed through even larger datasets \citep{uniGradIcon}. Second, while segmentation labels enhance performance in brain MRI \citep{jena2024deep}, these methods underperform on unlabeled data (e.g., DIR-Lab lung CT \citep{castillo2009framework}), trailing classical \citep{vishnevskiy2016isotropic} and INR-based \citep{vanharten2023robust} approaches. Finally, the CNNs and ViT by design typically cannot infer on several times higher or lower resolution without retraining, a limitation  partially resolved by resolution-robust neural operators \citep{drozdov2024fnoreg} at the cost of slightly reduced~accuracy.

To the best of our knowledge, there is no established universal SOTA method for medical image registration as a whole. While a few recent foundation-style approaches aim to build general-purpose DIR models \citep{uniGradIcon}, state-of-the-art performance is still defined on a per-dataset basis—for example, the leading results on DIR-Lab \citep{vishnevskiy2016isotropic}, OASIS \citep{chen2022transmorph,jena2024deep}, and ACDC \citep{meng2024correlation}.

\textbf{INRs for Image Registration.} 
In INR-based registration, the deformation field $\Phi$ is modeled as a continuous neural function over the spatial domain $\Omega$, fundamentally differing from discrete image-to-image approaches. While INRs require optimizing a new network for each image pair, making it more computationally demanding than a single forward pass through a pretrained CNN or transformer, it offers significant advantages. First, by eliminating the need for pretraining on large dataset, the INR-based methods require significantly less computational resources. Second, pair-specific registration algorithms also inherently avoid domain shift issues. Finally, the continuous nature of INRs makes them resolution independent, allowing seamless application to data at arbitrary resolutions without architectural modifications.

INR-based medical image registration was pioneered by \citep{wolterink2022implicit} using SIRENs for lung CT, achieving performance competitive with both classical and learning-based methods.  Subsequent work extended this to brain MRI with alternative activations and cycle consistency \citep{byra2023exploring}, and later to robust bidirectional frameworks with uncertainty quantification~\citep{vanharten2023robust}. Another pairwise optimization approach (but not explicitly INR) reformulated registration via neural ODEs in NODEO \citep{wu2021nodeo}. Most recently, SINR \citep{sideri2024sinr} hybridized B-spline FFDs with~INRs.

As demonstrated in our experiments below, current methods struggle to achieve an optimal balance between essential DIR requirements, including registration accuracy and deformation smoothness, as well as INR-specific constraints arising from the need of per-case network training, such as computational efficiency, memory demands, and seed-dependent stability.

\textbf{KANs.} 
Modern deep learning relies heavily on MLPs, known for their universal approximation capability \citep{cybenko1989approximation, hornik1989multilayer}.
The Kolmogorov--Arnold approximation theorem for multivariate continuous functions \citep{kolmogorov1957representation, arnold1957functions} offers an alternative: any continuous $f: [0,1]^n \to \mathbb{R}$ can be expressed as
\begin{equation} \label{eq6}
f(x_1, x_2, \ldots, x_n) = \sum_{q=1}^{2n+1} \Phi_q\left( \sum_{p=1}^{n} \phi_{q,p}(x_p) \right),
\end{equation}
where $\phi_{q,p}: [0,1] \to \mathbb{R}$ and $\Phi_q: \mathbb{R} \to \mathbb{R}$ are continuous univariate functions.

The authors of \citep{liu2024kan} introduced shallow and deep Kolmogorov--Arnold Networks (KANs) based on (\ref{eq6}), where each $\phi_{q,p}$ is implemented as spline-based univariate functions. While theoretically universal, shallow KANs with limited basis functions face convergence issues. Deep KAN architectures with multiple layers address this issue.

Recent work has explored KANs as INRs across domains, including the adaptation of KAN layers for standard INR tasks (image fitting and occupancy volumes) and the development of Fourier-KAN for representing neural sound fields \citep{mehrabian2024implicit, li2025representing}. Several studies \citep{Liu2024KAN2, faroughi2025neural} have demonstrated that KANs outperform MLPs in accurately approximating functions of a physical or mathematical nature, like special functions or solutions to partial differential equations.  Image registration can be formulated as finding the solution to a Euler--Lagrange system of a specific target functional \citep{drozdov2024fnoreg}, providing strong justification for using KANs in this context.

\section{Materials and Methods}

\subsection{Problem Formulation}
The DIR problem can be considered as finding the deformation field $\Phi (x) = x + U(x)$ that minimizes the following objective:
\begin{equation} \label{reg_task}
\hat{\Phi} = \argmin_{\Phi} \mathcal{L}_{\text{data}}(M \circ \Phi, \: F) + \mathcal{L}_{\text{reg}}(\Phi),
\end{equation}
where $\mathcal{L}_{\text{data}}$ is a similarity measure between the fixed image $F$ and the warped moving image $M \circ \Phi$, and $\mathcal{L}_{\text{reg}}$ is a regularization term imposing constraints on the deformation field $\Phi$. We assume that the fixed and moving images are defined on the same domain, i.e.,
$M, F : \Omega \subset [-1, 1]^n \to \mathbb{R}$.

\subsection{KAN-IDIR Architecture}
We adopt the Chebyshev KAN \citep{sidharth2024chebyshev} as the architecture of INR. This network utilizes Chebyshev polynomials with a fixed grid as basis functions, in contrast to B-splines with an adaptive grid used in the original KAN \citep{liu2024kan}. The Chebyshev polynomial of the  $n^{\text{th}}$ order is defined analytically as
\begin{equation}
T_n(x) = \cos(n \cdot \arccos (x)), \: x \in [-1, 1].
\end{equation}

Using this definition, we can perform efficient vectorized computations of $T_n$ values, thus increasing the network's speed in comparison to the original B-spline representation.
Forward pass through the Chebyshev KAN's layer can be mathematically defined as~follows:
\begin{equation} \label{eq_poly}
\left[\mathbf{y}(\mathbf{x})\right]_{b,o} = \sum_{i=1}^{N_{in}} \sum_{d \in \mathcal{D}} T_{d}(\mathbf{x}_{b,i} ) \cdot \mathbf{C}_{i,o,d},
\end{equation}
 where $b$ indexes the samples in the data batch, $i$ indexes the input dimension, $o$ indexes the output dimension, and $d$ indexes the polynomial degree. Here, $\mathbf{x}$ represents the input tensor, $\mathbf{y}$ the output tensor, $\mathbf{C}$ denotes the tensor of learnable coefficients, and $\mathcal{D} \subset \mathbb{N}_0$ is a finite set of all basis polynomial degrees, which may be shared across layers or vary between them.
 
 We use normalization with the hyperbolic tangent function (tanh) to map input tensor values in $[-1, 1]$ interval before passing into Chebyshev polynomials. To stabilize the training process, we follow the approach of the original KAN model and incorporate a learnable skip-connection in each layer. More precisely, the forward pass through the $m^{\text{th}}$ layer of our network is determined by the following expression:
\begin{equation} \label{layer_eq}
L_{m}(\mathbf{x}) = W_m b(\mathbf{x}) +  \mathbf{y}_m\left(\mathbf{\widetilde{x}}\right), 
\end{equation}
where $\mathbf{x}$ is the input tensor, $\mathbf{\widetilde{x}}$ denotes the inputs after tanh normalization, $\mathbf{y}_m(\cdot)$ represents the nonlinear part as defined in Equation (\ref{eq_poly}) and $W_m$ is the learnable matrix of linear transformation. We have chosen $b(\mathbf{x}) = \text{silu}(\mathbf{x}) = x/(1 + e^{-x})$ as an activation function for linear skip-connection. 

Using the definitions above, the architecture of our model can be written as the composition of $N$ layers from Equation (\ref{layer_eq}):
\begin{equation}
 \Phi(\mathbf{x}) = (L_{N-1} \: \circ \: L_{N-2}  \: \circ \: \dots \:  \: L_{0})(\mathbf{x}), \: \mathbf{x} \in \Omega \: .
\end{equation}

\subsection{Randomized Basis Sampling}
Sparse approximations were among the most prominent concepts in signal processing and pattern recognition during the pre-deep learning era \citep{zhang2015survey}. The core idea of these algorithms is that a target signal can be approximately represented by a sparse linear combination of prototype signals from an overcomplete dictionary \citep{mallat1993matching, aharon2006ksvd}. Inspired by the successful application of sparse representations in the past, we explored whether adopting a similar approach for KANs could enhance the quality and efficiency when applying to the DIR task. In the case of the KAN model, the set of basis functions serves as an overcomplete dictionary, and our goal is to utilize only a few of them during each forward pass of data through the network. Thus, two strategies can be considered: either selecting a \textit{fixed 
} sparse set of basis functions for each layer before training or learning the optimal selection \textit{dynamically} during network optimization. Based on these two strategies together with classical sequential selection of basis functions, we evaluate three distinct models in our subsequent work:

\vspace{3pt}
\noindent \textbf{KAN-IDIR:} 
 This model employs the classical basis selection scheme with a fixed maximum degree $D$. For each layer, the basis functions are polynomials with degrees from $0$ to $D$, i.e., $\mathcal{D} = \{0, 1, \dots, D\}$. The set $\mathcal{D}$ of polynomial degree indices is shared across all layers.

\noindent \textbf{RandKAN-IDIR}: This model implements basis selection with randomized sampling. Let $\mathcal{D}_m$ denote the set of degrees of the basis polynomials for the $m^{\text{th}}$ layer. It is constructed by including the constant term (degree 0) and combining it with a random selection of higher degrees: $\mathcal{D}_m = \{0\} \cup \mathcal{S}_m$, where $\mathcal{S}_m$ is a simple random sample without replacement of size $k$ from the set $\{1, 2, \dots, K\}$, independently for each layer $m$. Despite the reduced basis set, experiments (Section \ref{sec:experiments}) show that the RandKAN-IDIR model maintains competitive registration accuracy while improving efficiency, given an appropriate choice of $k$ and $K$. Here, $k$ controls the number of active non-constant basis functions per layer, while $K$ defines the size of the candidate Chebyshev dictionary. The case $k = K$ corresponds to the standard consecutive low-degree selection, whereas $K > k$ gives a sparse non-consecutive subset from a broader set of polynomial degrees. Our hypothesis is that this provides access to selected higher-degree components without increasing the number of active basis functions. This behavior is analyzed empirically in Section~\ref{sec:ablation_k_K}.

\noindent \textbf{A-KAN-IDIR}: This variant implements a learned sparse basis selection mechanism using adaptive basis weights and a Noisy Top-k procedure. It is used as a learned sparse-mask alternative to RandKAN-IDIR (see Appendix \ref{sec:akanidir} 
 for details). Experiments show no quality improvement over RandKAN-IDIR's random selection, while being computationally slower, so we exclude this model from the evaluations.

\subsection{Training Objective}

Our loss function combines similarity and regularization terms (Equation (\ref{reg_task})). For the similarity term, we use negative normalized cross-correlation (NCC):
\begin{flalign}
& \mathcal{L}_{\text{data}}(I, J) \!=\! -NCC(I, J) \notag = \frac{-\sum_{\mathbf{p} \in \Omega_p} \left(I(\mathbf{p}) - \bar{I}\right)\left(J(\mathbf{p}) - \bar{J}\right)}{\sqrt{\sum_{\mathbf{p} \in \Omega_p} \left(I(\mathbf{p}) - \bar{I}\right)^2 \sum_{\mathbf{p} \in \Omega_p} \left(J(\mathbf{p}) - \bar{J}\right)^2}},
\end{flalign}
where $\Omega_p \subset \Omega$ is the current batch of voxel coordinates, and $\bar{I},\bar{J} \in \mathbb{R}$ are the mean intensities of corresponding images over $\Omega_p$.

The regularization term $\mathcal{L}_{\text{reg}}$ combines two components: a smoothness term $\mathcal{L}_{\text{smooth}}$ and a folding prevention term $\mathcal{L}_{\text{Jdet}}$, weighted by parameters $\lambda$ and $\gamma$ respectively. To enforce physically plausible deformations, we employ total variation (TV) regularization~\citep{Rudin1992NonlinearTV}, which penalizes the L1 norm of the displacement field gradient:
\begin{equation}
\mathcal{L}_{\text{smooth}} = \frac{1}{|\Omega_p|} \sum_{\mathbf{p} \in \Omega_p}||\nabla U(\mathbf{p})||_1.
\end{equation}

While 
total variation (TV) regularization effectively promotes smooth deformation fields, it may still permit folded voxels, i.e., voxels with locally non-diffeomorphic transformations characterized by negative Jacobian determinants ($|J_{\Phi}| < 0$). To address this, we incorporate a Jacobian determinant regularization term \citep{mok2020fast} that actively penalizes folding regions:
\begin{equation} \label{jdet_loss}
\mathcal{L}_{\text{Jdet}} = \frac{1}{|\Omega_p|} \sum_{\mathbf{p} \in \Omega_p} \sigma(-|J_{\Phi}(\mathbf{p})| + \epsilon),
\end{equation}
where $\sigma(x) = \max(0, x)$ is the ReLU function and $\epsilon > 0$ is a small constant added for overcorrection \citep{wu2021nodeo}. The complete training objective combines these regularization terms with the data similarity term:
\begin{equation} \label{tr_obj}
\mathcal{L} = \mathcal{L}_{data} + \lambda \mathcal{L}_{\text{smooth}} + \gamma \mathcal{L}_{\text{Jdet}}.
\end{equation}

\begin{figure}[H]
\begin{adjustwidth}{-\extralength}{0cm}
\centering
\includegraphics[width=0.8\fulllength]{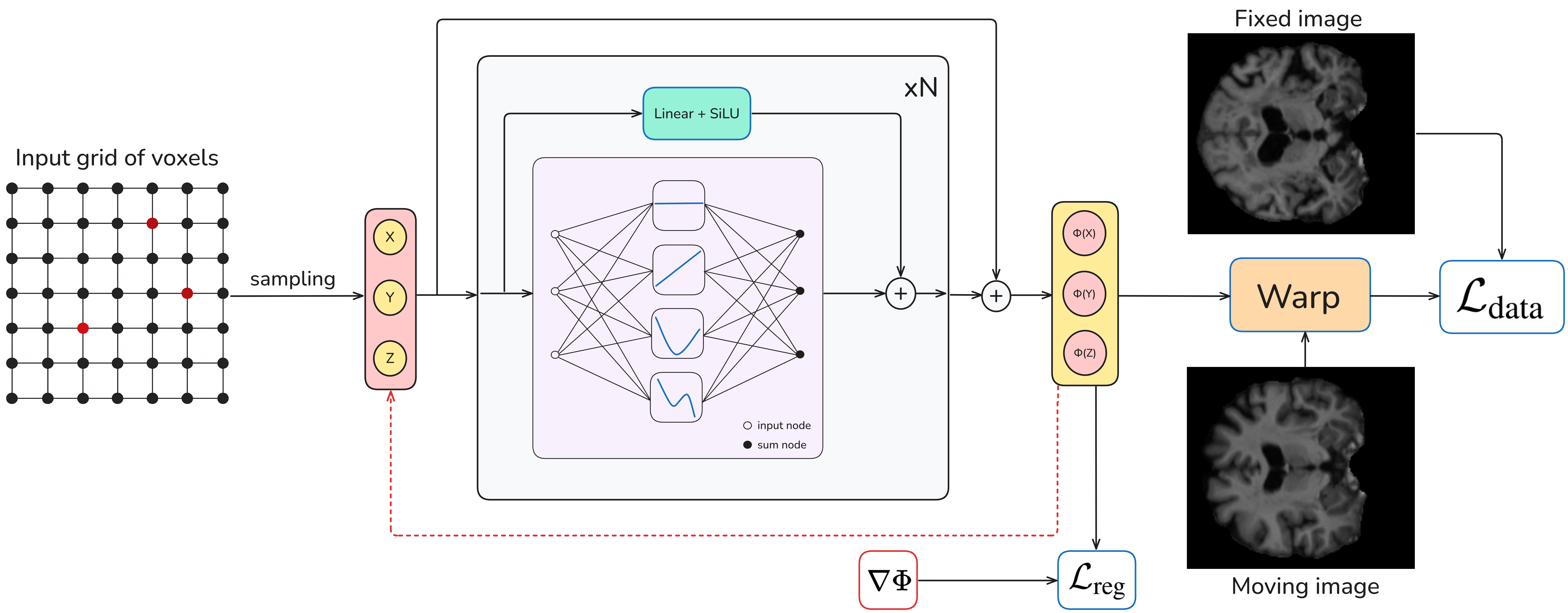}
\end{adjustwidth}
\caption{The 
 overall pipeline of the proposed method for deformable image registration. Voxels are sampled from the input volume to serve as input to the KAN-IDIR or RandKAN-IDIR models. The network outputs a deformation field, which is used to warp the moving image, enabling calculation of the data similarity loss. For regularization, the gradient of the deformation field is computed using automatic differentiation.}
\label{pipeline}
\end{figure}

The overall pipeline of the proposed method for deformable image registration is shown in Figure \ref{pipeline}.

\subsection{Datasets and Preprocessing}

We evaluated our approach using three established DIR datasets spanning three organs and two imaging modalities.

\textbf{Lung CT:}
For the lung CT image registration task, we utilized the DIR-Lab dataset  \citep{castillo2009framework}, which includes 4D CT scans from ten patients, capturing inspiration and expiration phases for registration. This dataset is widely adopted for evaluating DIR methods due to the complex motion compensation challenge posed by the interplay of cardiac and respiratory movements, which significantly exceed the scale of small lung structures. Each image pair is annotated with 300 human-identified landmarks corresponding to stable anatomical features. The images vary in size, with in-plane resolutions ranging from $256 \times 256$ to $512 \times 512$ pixels and Z-axis resolutions spanning 94 to 136 pixels.

\textbf{Brain MRI:} 
OASIS-1 dataset \citep{marcus2007open} consists of brain MRI scans from 414 subjects, each accompanied by segmentations of 35 key anatomical regions. For our experiments, we employed a preprocessed version of the 3D OASIS-1 dataset \citep{hoopes2022learning}, where all 414 MRI scans were affinely aligned and cropped to a resolution of $160 \times 192 \times 224$. We selected the final 50 subjects to create 49 test pairs, adhering to the pairing rules outlined in the Learn2Reg challenge \citep{hering2022learn2reg} for evaluating the proposed approach.

\textbf{Cardiac MRI:} 
We utilized the ACDC dataset \citep{bernard2018deep} for cardiac MRI registration, which includes 150 image pairs acquired during the end-diastolic (ED) and end-systolic (ES) phases, accompanied by segmentation labels for the left ventricular cavity, right ventricular cavity, and myocardium. The ED is defined as the first frame when the mitral valve closes, and the ES is defined as the first frame when the aortic valve closes. For evaluation, we used a test subset of 50 image pairs from the ACDC dataset, registered bidirectionally (ED-to-ES and ES-to-ED), yielding 100 test image pairs. All images were preprocessed and cropped in accordance with the methodology outlined in CorrMLP \citep{meng2024correlation}.

\subsection{Evaluation Metrics}

For all datasets, we evaluated four quantitative metrics:

\begin{itemize}
   \item Registration accuracy, measured by Target Registration Error (TRE) \citep{castillo2009framework} for DIR-Lab and by Dice score (DSC) and the 95\textsuperscript{th} percentile Hausdorff distance (HD95) for OASIS-1 and ACDC. TRE is the mean Euclidean distance between landmarks in the fixed image and the corresponding landmarks from the warped moving image. The Dice score for a pair of images with $N$ anatomical structures is calculated as
    \begin{equation}
    \text{DSC}(\mathbf{S}_{F}, \mathbf{S}_{M \circ \Phi}) = \frac{2}{N}\sum\limits_{n=1}^{N}\frac{|\mathbf{S}^{n}_{F} \cap \mathbf{S}^{n}_{M \circ \Phi}|}{|\mathbf{S}^{n}_{F}| + |\mathbf{S}^{n}_{M \circ \Phi}|} ,
    \end{equation}
    
    where $\mathbf{S}^{n}_{F}$ and $\mathbf{S}^{n}_{M \circ \Phi}$ are the segmentation labels of the $n$-th structure in the fixed and warped moving images, respectively.
    The HD95 metric is computed similarly to DSC, by averaging the HD95 values over individual segmentations. More details are provided in the Appendix \ref{sec:hd95}. 

   \item Deformation regularity, quantified as the percentage of folded voxels (NJD):
    \begin{equation}
    \text{NJD} = \frac{\left| \{ \mathbf{p} : |J_{\Phi}(\mathbf{p})| < 0 \}\right|}{\text{total voxels}} \times 100\%,
    \end{equation}
    {Here,} $|J_{\Phi}(\mathbf{p})|$ denotes the determinant of the Jacobian matrix of the transformation $\Phi$.
    For additional folding-severity analysis, we report the mean magnitude of negative Jacobian determinants,
    \begin{equation}
    \mathrm{MeanNegJ} = \mathbb{E}\left[-|J_{\Phi}(\mathbf{p})| \mid |J_{\Phi}(\mathbf{p})| < 0\right],
    \end{equation}
    and the percentage of voxels with severe foldings,
    \begin{equation}
    \mathrm{SevereJ}_{0.1} = \frac{\left| \{ \mathbf{p}: |J_{\Phi}(\mathbf{p})| < -0.1 \}\right|}{|\Omega|}\times100\%.
    \end{equation}
    {These} metrics distinguish the spatial extent of foldings from their local severity.
    \item GPU memory consumption during the registration process.  
    \item Algorithm runtime, measured as the total wall-clock time for registering an image pair.
\end{itemize} Experiments ran on an Intel Xeon Gold 6226R CPU/NVIDIA A6000 server under exclusive load to ensure runtime precision measurement. For INR-based methods, metrics were averaged over ten random seeds (mean $\pm$ std), following \citep{vanharten2023robust,wolterink2022implicit}.

\setlength{\tabcolsep}{1mm}

\subsection{Baselines and Implementation Details}
We implemented our method using the PyTorch 2.3.0 
 library \citep{paszke2019pytorch}. All KAN-based models consist of two hidden layers, each with 70 neurons. This configuration was chosen to approximately match the number of parameters in our models to that of classical INR-based registration approaches \citep{wolterink2022implicit, vanharten2023robust}. For the final configuration, we set $\lambda = 0.4$, $\gamma = 15$ and $\epsilon = 0.1$ in Equations (\ref{jdet_loss}) and (\ref{tr_obj}) unless otherwise specified; the values of $\lambda$ and $\gamma$ were selected from the Pareto frontier in the loss-weight ablation (Section~\ref{sec:ablation_gamma_lambda}) 
 to balance registration accuracy and deformation regularity. The Adam optimizer \citep{kingma2014adam} was employed for model training with an initial learning rate of $10^{-4}$. The learning rate remained constant for the first 50\% of the iterations, after which it decreased following a cosine annealing schedule. Each model was trained for a total of 1500 iterations, with 10,000 points randomly sampled from the domain of interest on each iteration. The domain of interest is defined by a segmentation mask that separates the organ from the background. For the DIR-Lab dataset, we followed the approach from \citep{wolterink2022implicit} and used the pretrained segmentation model to obtain the lung mask \citep{Hofmanninger2020AutomaticLungSegmentation}. The OASIS-1 dataset provides reference masks for anatomical segments, enabling us to use the union of all non-background segmentations as the resulting mask. For the ACDC dataset, we used all image voxels for training. For comprehensive baseline implementation details we refer readers to the Appendix \ref{sec:baseline_impl}. 

\section{Results}
\label{sec:experiments}
Our evaluation compares against: neural ODE-based NODEO \citep{wu2021nodeo}; INR-based methods (IDIR \citep{wolterink2022implicit}, ccIDIR \citep{vanharten2023robust}, and SINR \citep{sideri2024sinr}); dataset-specific SOTA (pTV \citep{vishnevskiy2016isotropic} for DIR-Lab, CorrMLP \citep{meng2024correlation} for ACDC); and established DIR baselines (VoxelMorph \citep{balakrishnan2019tmi}, TransMorph \citep{chen2022transmorph} for~OASIS-1).

We denote KAN-IDIR ($D$) as the model configuration with a maximum degree of $D$. For the RandKAN-IDIR model, we use a configuration with \(k = 12\) and \(K = 84\) in all experiments unless otherwise specified.

\subsection{DIR-Lab Results}

Table~\ref{dirlab_table} presents the experimental results obtained on the DIR-Lab dataset. KAN-IDIR and RandKAN-IDIR achieve the second and third best TRE, respectively, outperformed only by the classical pTV method, which remains SOTA for this dataset against all classical and DL-based approaches. However, our methods are significantly faster: RandKAN-IDIR registers pairs in under one minute versus pTV's several minutes. Additionally, the performance difference between RandKAN-IDIR and KAN-IDIR is minimal, with only 0.01~mm average TRE gap,  while RandKAN-IDIR achieves approximately 30\% faster execution (43.1 s versus 63.3 s).
Notably, while NODEO and SINR perform competitively on other modalities, they show poor lung CT performance with TRE $\approx4$ times higher than our~methods.

\begin{figure}[H]
\begin{adjustwidth}{-\extralength}{0cm}
\centering
\includegraphics[width=0.9\linewidth]{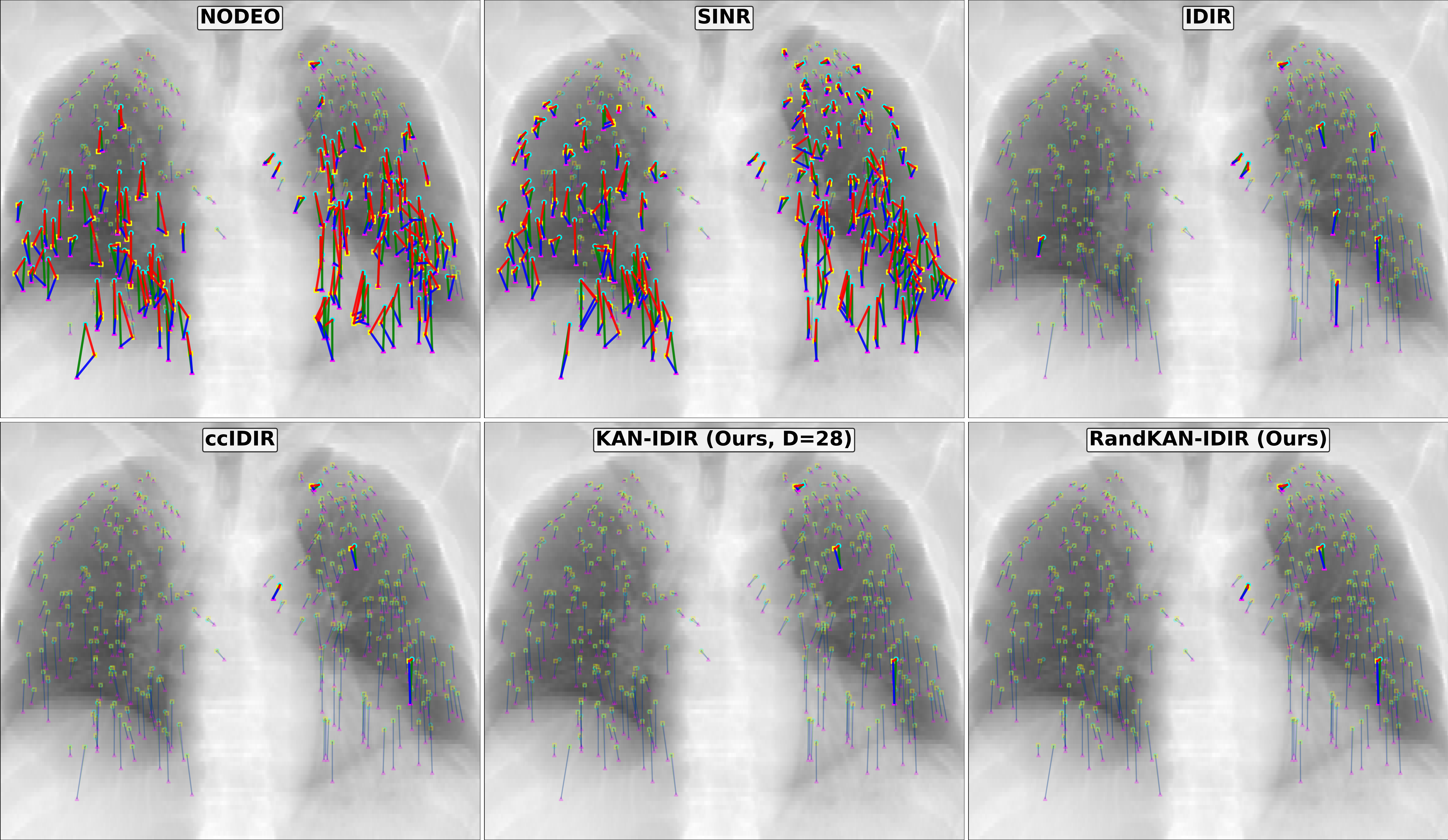}
\end{adjustwidth}
\caption{Landmark alignment in DIR-Lab Dataset (Case 7) for fixed, moving, and moved images. Displacements are shown as: blue (method result, moved-to-fixed), green (ground truth, moving-to-fixed), and red (error, moved-to-fixed). Landmarks with errors exceeding 3 mm are highlighted in bold, while others are displayed as transparent. The results reflect each method's optimal seed (minimum mean TRE across 10 runs). Proposed KAN-IDIR and RandKAN-IDIR achieve the fewest outliers. The best seed-dependent variability of the proposed methods is demonstrated in Figure \ref{dirlab_results_max_seed} in Appendix 
 using worst-case (maximum mean TRE) results.}
\label{dirlab_results}
\end{figure}

\begin{figure}[H]
\includegraphics[width=0.99\columnwidth]{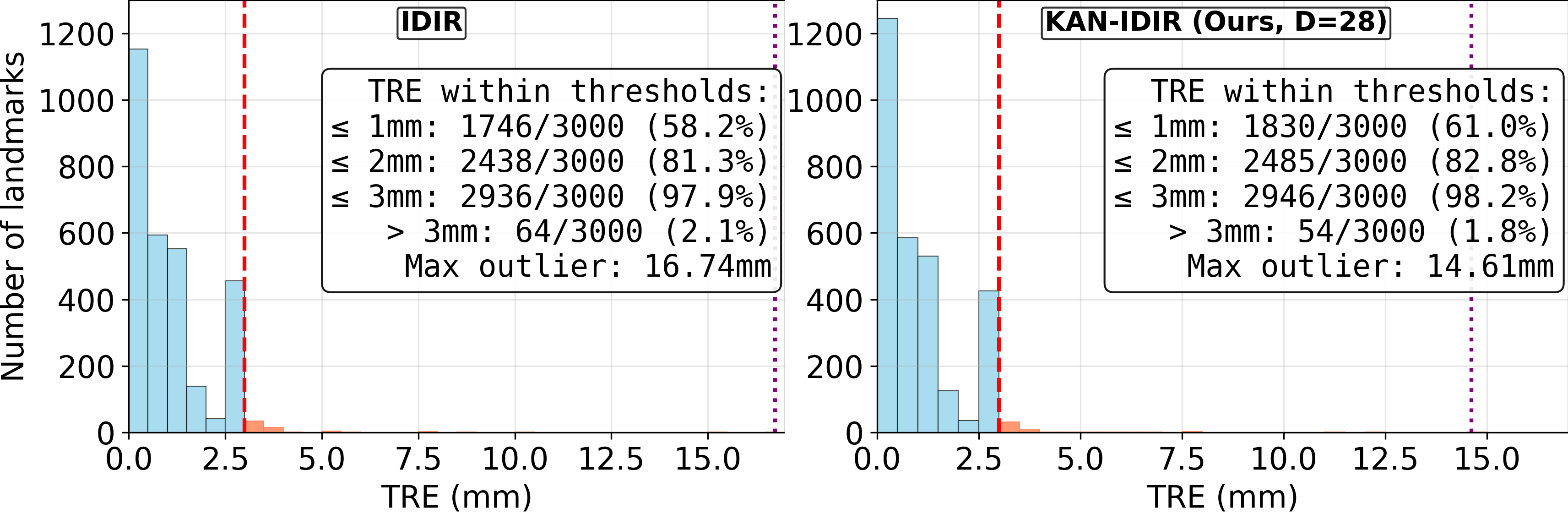}\\
\includegraphics[width=0.99\columnwidth]{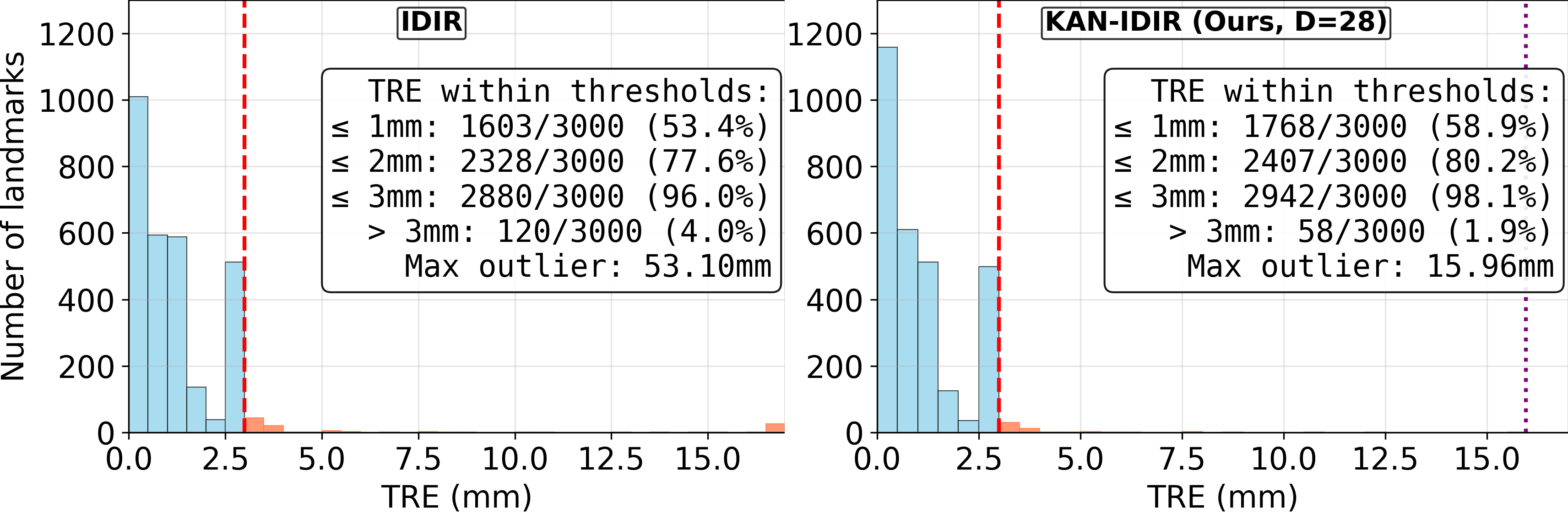}
\caption{Target registration error (TRE) distributions for IDIR and proposed KAN-IDIR on DIR-Lab dataset. Upper two histograms show results for each method’s best seed (lowest mean TRE), and lower two histograms for the worst seed (highest mean TRE) out of 10 runs. The dashed red line marks the 3 mm threshold; the dotted purple line marks maximum outlier. KAN-IDIR produces fewer >3 mm outliers under the best-performing seed (54, compared to 64 for IDIR) and also exhibits substantially lower sensitivity to the choice of seed. Specifically, the outlier count for IDIR increases from 64 (best seed) to 120 (worst seed), a difference of 56, whereas KAN-IDIR increases only from 54 to 58, a difference of 4. This large contrast indicates that KAN-IDIR is considerably more robust to initialization.}
\label{hist_results}
\end{figure}

Figure~\ref{dirlab_results} visualizes the overlay of the human-defined landmarks used to calculate TRE, showing the qualitative behavior of the evaluated INR-based methods. Figure~\ref{hist_results} depicts the distribution of registration errors across all landmark pairs in the DIR-Lab dataset, while Table~\ref{tab:outlier_results} presents the percentage of errors within specified thresholds. It demonstrates that our methods not only achieve superior average registration quality but also exhibit improved quality in terms of error distribution and seed-dependent variability. Under the best seed, KAN-IDIR produces fewer $>$3 mm outliers than IDIR (54 versus 64). Under the worst seed, the outlier count for IDIR increases to 120, whereas KAN-IDIR remains nearly unchanged at 58, indicating substantially higher robustness to random initialization.

\begin{table}[H]
\caption{Evaluation 
 of baselines and proposed methods (marked with *) 
 on the DIR-Lab dataset. Values in parentheses denote standard deviations, and---indicates that the standard deviation was not reported. The best result is in \textbf{bold}, and the second-best is in \textit{italics}. Our models achieve near state-of-the-art performance in lung CT registration with the lowest execution times.}\vspace{-3pt}
\label{dirlab_table}
\begin{adjustwidth}{-\extralength}{0cm}
\centering
{%
\setlength{\tabcolsep}{1mm}
\renewcommand{\arraystretch}{1.2}
\fontsize{9pt}{11pt}\selectfont
\begin{tabularx}{\fulllength}{C|c|cccccc}
\noalign{\hrule height 1pt} 
\textbf{Methods} & \textbf{pTV} & \textbf{NODEO} & \textbf{SINR} & \textbf{IDIR} & \textbf{ccIDIR} & \textbf{KAN-IDIR (28) *} & \textbf{RandKAN-IDIR *} \\ \hline
Average 
 TRE, mm $\downarrow$ 
 & \textbf{0.95 (1.15)} & 3.93 (6.07) & 4.63 (5.26) & 1.07 (1.10) & 1.04 (1.12) & \textit{0.98 (1.10)} & 0.99 (1.10) \\ \hline
NJD, \% $\downarrow$ & 0.6 (--) & \textbf{0.0003 (0.0003)} & 0.002 (0.002)& 0.002 (0.015) & 0.002 (0.008) & \textit{0.0006 (0.001)} & 0.006 (0.016) \\
Runtime, s $\downarrow$ & 442 & 583.5 & 91.6 & 260.8 & 92.6 & \textit{63.3} & \textbf{43.1} \\
VRAM, Gb $\downarrow$ & -- & 18.7 & 21.7 & 4.1 & \textbf{1.0} & 2.2 & \textit{1.4} \\ \noalign{\hrule height 1pt} 
\end{tabularx}
}
\end{adjustwidth}
\end{table}

\vspace{-6pt}

\vspace{-6pt}

\begin{table}[H]
\caption{Cumulative distribution of TRE for DIR-Lab data shown as percentages within thresholds ($\leq$1--3 mm) and for outliers (>3 mm) across pairwise DL methods. Proposed method names and best values in \textbf{bold}.}\vspace{-3pt}
\label{tab:outlier_results}
\setlength{\tabcolsep}{1mm}
\renewcommand{\arraystretch}{1.2}
\fontsize{9pt}{11pt}\selectfont
\centering
\begin{tabularx}{\textwidth}{L|CcCc}
\noalign{\hrule height 1pt} 
\textbf{Method} & \boldmath{$\leq$}\textbf{1 mm} \boldmath{$\uparrow$} & \boldmath{$\leq$}\textbf{2 mm} \boldmath{$\uparrow$} & \boldmath{$\leq$}\textbf{3 mm} \boldmath{$\uparrow$} & >\textbf{3 mm} \boldmath{$\downarrow$} \\
\hline
NODEO & 39.05\% & 57.23\% & 73.09\% & 26.91\% \\
SINR & 18.14\% & 34.56\% & 55.59\% & 44.41\% \\
IDIR & 56.51\% & 79.78\% & 97.52\% & 2.48\% \\
ccIDIR & 57.07\% & 79.97\% & 97.95\% & 2.05\% \\
\textbf{KAN-IDIR (28)} & \textit{60.13\%} & \textbf{81.69\%} & \textbf{98.14\%} & \textbf{1.86\%} \\
\textbf{RandKAN-IDIR} & \textbf{60.28\%} & \textit{81.43\%} & \textit{98.09\%} & \textit{1.91\%} \\
\noalign{\hrule height 1pt} 
\end{tabularx}
\end{table}

\subsection{OASIS-1 Results}

For the OASIS-1 dataset, we also include the learning-based methods such as VoxelMorph and TransMorph in the comparison. Table~\ref{tab:oasis_results} shows the results of experiments on the OASIS-1 dataset. TransMorph achieves the highest DSC overall, while KAN-IDIR and RandKAN-IDIR achieve the best mean DSC and HD95 among the evaluated INR-based methods. To assess whether the observed differences are statistically supported, we additionally performed one-sided paired Wilcoxon signed-rank tests on per-patient-pair mean Dice scores. KAN-IDIR significantly outperformed SINR, IDIR, ccIDIR, and NODEO, and RandKAN-IDIR significantly outperformed SINR, IDIR, and ccIDIR, while its difference with NODEO was not statistically significant. At the same time, the absolute DSC differences between the strongest pairwise methods are small, so the OASIS-1 results should be interpreted together with runtime, memory consumption, deformation regularity, and training requirements. We discuss the latter trade-off explicitly in Section~\ref{sec:runtime_amortization}. Additional folding-severity metrics are provided in the Appendix (Table~\ref{tab:oasis_folding_severity}). The visual results for the OASIS-1 dataset are provided in the Appendix (Figure~\ref{fig:oasis_collage}).

\subsection{ACDC Results}

Table~\ref{tab:acdc_results} presents the experimental results for the ACDC dataset. Most methods exhibit similar performance levels, which we attribute to the relatively small number of voxels available for optimization, promoting a more stable training process. 
The proposed KAN-IDIR model with $D=8$ reaches the highest DSC among the compared methods and exceeds the reported in \citep{meng2024correlation} DSC, while also achieving substantially better deformation regularity (only 0.00002\% of foldings).
Similarly, RandKAN-IDIR provides competitive registration quality while exhibiting excellent deformation field regularity. The visual comparison for the ACDC dataset is presented in the Appendix (Figure \ref{fig:acdc_collage}).

\begin{table}[H]
\caption{Results 
 of experiments on the OASIS-1 dataset. Values in parentheses denote standard deviations. For learning-based methods, runtime and memory consumption are reported for both inference and training in the format inference/training. Our KAN-IDIR and RandKAN-IDIR achieve the highest mean registration quality among the evaluated INR-based methods, with nearly the fastest execution times and low VRAM consumption. Proposed method names marked with *, 
 best-in-class values in \textbf{bold}, second-best in \textit{italics}.}\vspace{-3pt}
\label{tab:oasis_results}
\begin{adjustwidth}{-\extralength}{0cm}
\centering
{%
\renewcommand{\arraystretch}{1.2}
\fontsize{9pt}{11pt}\selectfont
\begin{tabularx}{\fulllength}{C|cc|cccccc}
\noalign{\hrule height 1pt} 
\textbf{Methods} & \textbf{VM} & \textbf{TM} & \textbf{NODEO} & \textbf{SINR} & \textbf{IDIR} & \textbf{ccIDIR} & \textbf{KAN-IDIR (28) *} & \textbf{RandKAN-IDIR} * \\ \hline
DSC $\uparrow$ & \textit{0.795 (0.03)} & \textbf{0.811 (0.02)} & 0.790 (0.02) & 0.760 (0.03) & 0.762 (0.03) & 0.779 (0.02) & \textbf{0.793 (0.02)} & \textit{0.792 (0.02)} \\
HD95 $\downarrow$ & \textit{1.95 (0.34)} & \textbf{1.81 (0.30)} & 1.98 (0.38) & 2.24 (0.37) & 2.22 (0.45) & 2.19 (0.44) & \textbf{1.96 (0.32)} & \textit{1.97 (0.32)} \\
NJD, \% $\downarrow$ & \textit{0.530 (0.15)} & \textbf{0.526 (0.14)} & \textbf{0.011 (0.01)} & 2.004 (0.52) & \textit{0.041 (0.03)} & 0.345 (0.12) & 0.046 (0.03) & 0.043 (0.03) \\
Runtime $\downarrow$ & \textbf{0.1 s/2.1 days} & \textit{0.2 s/3.7 days} & 68.1 s & \textit{50.3 s} & 260.8 s & 92.6 s & 63.3 s & \textbf{43.1 s} \\
VRAM, Gb $\downarrow$ & \textbf{8.7/14.1} & \textit{19.8/22.0} & 2.7 & 9.0 & 4.4 & \textbf{1.0} & 2.2 & \textit{1.4} \\ \noalign{\hrule height 1pt} 
\end{tabularx}
}
\end{adjustwidth}
\end{table}
\vspace{-9pt}

\begin{table}[H]
\caption{Results 
 of experiments on the ACDC dataset. Values in parentheses denote standard deviations. Values marked with $\dagger$ are taken directly from \citep{meng2024correlation}. KAN-based models achieve competitive performance with superior deformation regularity. Proposed method names marked with *, 
 best values in \textbf{bold}, second-best in \textit{italics}.}\vspace{-3pt}
\label{tab:acdc_results}
\begin{adjustwidth}{-\extralength}{0cm}
\centering
{%
\setlength{\tabcolsep}{1mm}
\renewcommand{\arraystretch}{1.2}
\fontsize{8pt}{11pt}\selectfont
\begin{tabularx}{\fulllength}{c|c|ccccccc}

\noalign{\hrule height 1pt} 
\textbf{Methods} & \textbf{CorrMLP} & \textbf{NODEO} & \textbf{SINR} & \textbf{IDIR} & \textbf{ccIDIR} & \textbf{KAN-IDIR (8) *} & \textbf{KAN-IDIR (28) *} & \textbf{RandKAN-IDIR *} \\ \hline
DSC $\uparrow$ & 0.810 (0.08) & 0.795 (0.07) & \textit{0.811 (0.05)} & \textbf{0.814 (0.06)} & 0.792 (0.06) & \textbf{0.814 (0.06)} & 0.809 (0.06) & 0.810 (0.06) \\
HD95 $\downarrow$ & -- & 7.87 (3.17) & \textbf{6.75 (3.11)} & \textit{7.38 (3.38)} & 8.23 (3.15) & 7.39 (3.17) & 7.87 (3.25) & 7.88 (3.28) \\
NJD, \% $\downarrow$ & 0.389 (0.143) & 0.011 (0.045) & 0.002 (0.002) & 0.111 (0.166) & 0.051 (0.066) & \textbf{2 \boldmath{$\times~10^{-5}$} (2 \boldmath{$\times~10^{-4}$}) }& 5 $\times~10^{-5}$ \mbox{(4 $\times~10^{-4}$)} &\textit{ 4 $\times~\text{10}^{-\text{5}}$ (2 $\times~\text{10}^{-\text{4}}$)} \\
Runtime, s $\downarrow$ & 0.07 $\dagger$ & \textit{19.6} & \textbf{13.9} & 260.8 & 92.6 & 40.6 & 63.3 & 43.1
\\
VRAM, Gb $\downarrow$ & - & \textbf{0.7} & 2.6 & 4.4 & \textit{1.0} & 1.1 & 2.2 & 1.4 \\ \noalign{\hrule height 1pt} 
\end{tabularx}
}
\end{adjustwidth}
\end{table}

For additional context, we also compare against uniGradICON \citep{uniGradIcon}, a recent foundation-style registration model trained on large-scale data, in the Appendix (Table~\ref{tab:unigrad_results}). This comparison places the proposed pairwise optimization approach relative to general-purpose pretrained registration: KAN-IDIR is comparable to uniGradICON on OASIS-1 and achieves higher DSC and lower HD95 on ACDC.

\subsection{Runtime Amortization}
\label{sec:runtime_amortization}
Inference-only runtimes for learning-based methods should be interpreted together with training amortization. On OASIS-1, VoxelMorph and TransMorph require 2.1 and 3.7 days for training, respectively. Including inference on the 49 test pairs, the total time remains approximately 2.1 and 3.7 days, whereas KAN-IDIR and RandKAN-IDIR process the same pairs in 51.7 and 35.2 min. These training costs are amortized only after approximately 2871/4220 registrations for VoxelMorph relative to KAN-IDIR/RandKAN-IDIR and 5066/7452 registrations for TransMorph. On DIR-Lab, which contains only 10 image pairs, training a dataset-level model from scratch is not practical, while KAN-IDIR and RandKAN-IDIR process all pairs in 10.6 and 7.2 min. On ACDC, the 100 bidirectional test registrations require 67.7 min with KAN-IDIR ($D=8$) and 71.8 min with RandKAN-IDIR. Thus, feed-forward models are preferable when a sufficiently large training cohort and many registrations are available, while pairwise methods are most useful for small or data-scarce studies where training a dataset-level model is unreliable or unnecessary.

\subsection{Convergence Speed}
\begin{figure}[H]
\includegraphics[width=0.95\columnwidth]{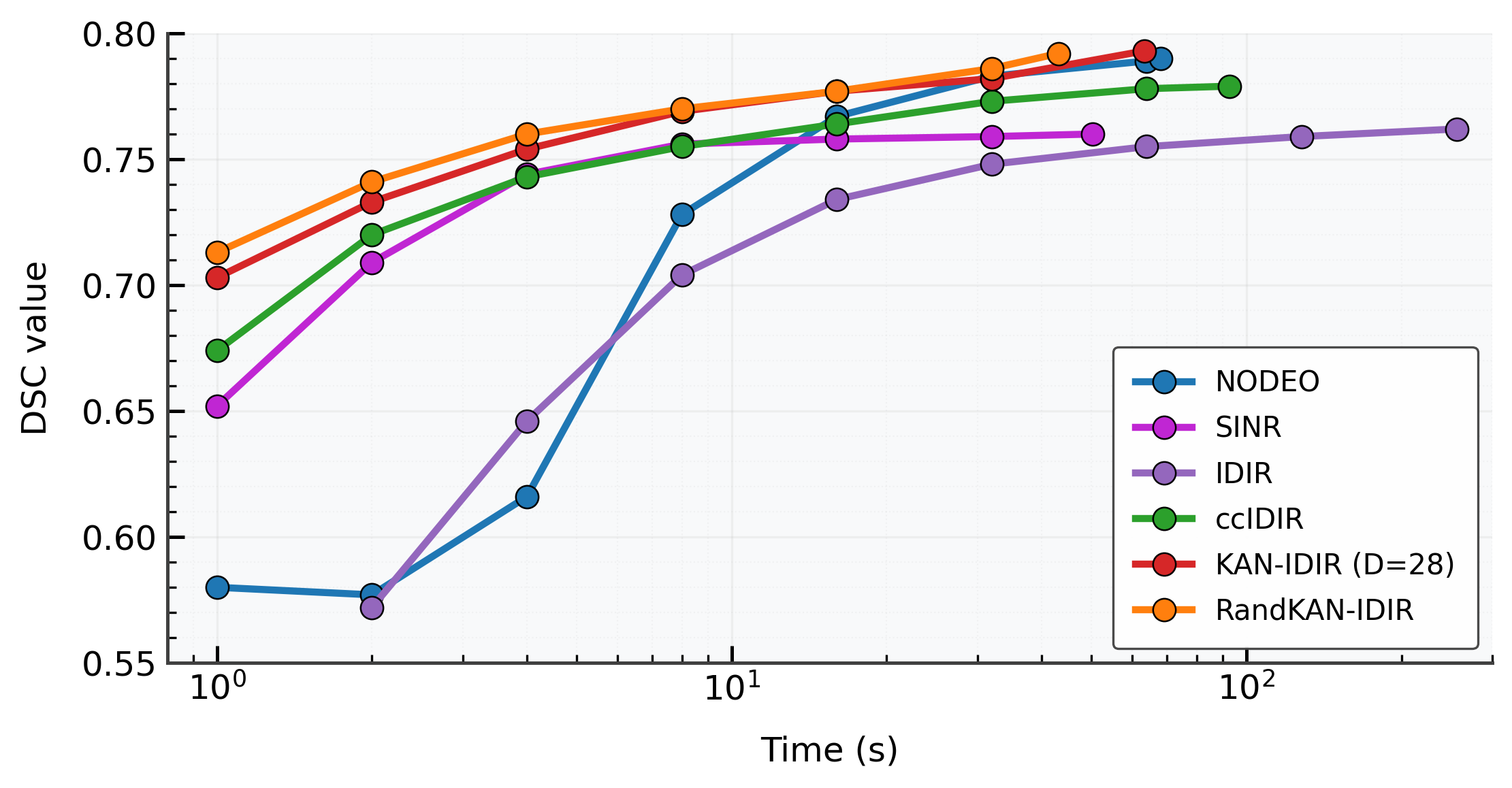}\vspace{-6pt}
\caption{Dependency of the Dice Similarity Coefficient (DSC) on optimization time for different pairwise registration methods on the OASIS-1 dataset.}
\label{fig:conv_speed}
\end{figure}

Figure \ref{fig:conv_speed} illustrates the convergence behavior of different pairwise registration methods on the OASIS-1 dataset, measured by DSC values at consecutive time points during optimization. KAN-IDIR and RandKAN-IDIR demonstrate the fastest convergence, outperforming all other methods except NODEO after only 20 s of optimization. During the initial phase of training, our models achieve 10–20\% higher performance than NODEO, indicating stronger initial alignment and faster convergence toward the optimal deformation.

\subsection{Ablation Studies} \label{sec:ablation}

\subsubsection{Ablation for Maximum Polynomial Degree} In Table~\ref{tab:ablation_dirlab}, we evaluate the performance of the KAN-IDIR ($D$) model on the DIR-Lab dataset across various values of $D$. The results indicate that a value of $D = 16$ is sufficient for KAN-IDIR to achieve state-of-the-art performance among INR-based methods on the DIR-Lab dataset. For $D = 28$, the performance reaches a plateau, with minimal further improvements observed for higher degrees.

\begin{table}[H]
\caption{Ablation 
 study of basis functions number $D$ 
 in KAN-IDIR model on the DIR-Lab dataset. KAN-IDIR model with $D = 28$ delivers best registration quality. Best in \textbf{bold}.}\vspace{-3pt}
\label{tab:ablation_dirlab}
\centering
\setlength{\tabcolsep}{1mm}
\renewcommand{\arraystretch}{1.2}
\begin{tabularx}{\textwidth}{C|CCCCCCCC}
\noalign{\hrule height 1pt} 
\textbf{$D$} &  \textbf{8} & \textbf{12} & \textbf{16} & \textbf{20} & \textbf{24} & \textbf{28} & \textbf{36} \\ \hline
Avg. TRE, mm $\downarrow$ & 1.11 & 1.04 & 1.01 & \textit{0.99} & \textit{0.99} & \textbf{0.98} & \textbf{0.98} \\ \noalign{\hrule height 1pt} 
\end{tabularx}
\end{table}

\subsubsection{Ablation of Loss Function Weights}
\label{sec:ablation_gamma_lambda}
\begin{figure}[H]
        \includegraphics[width=0.95\linewidth]{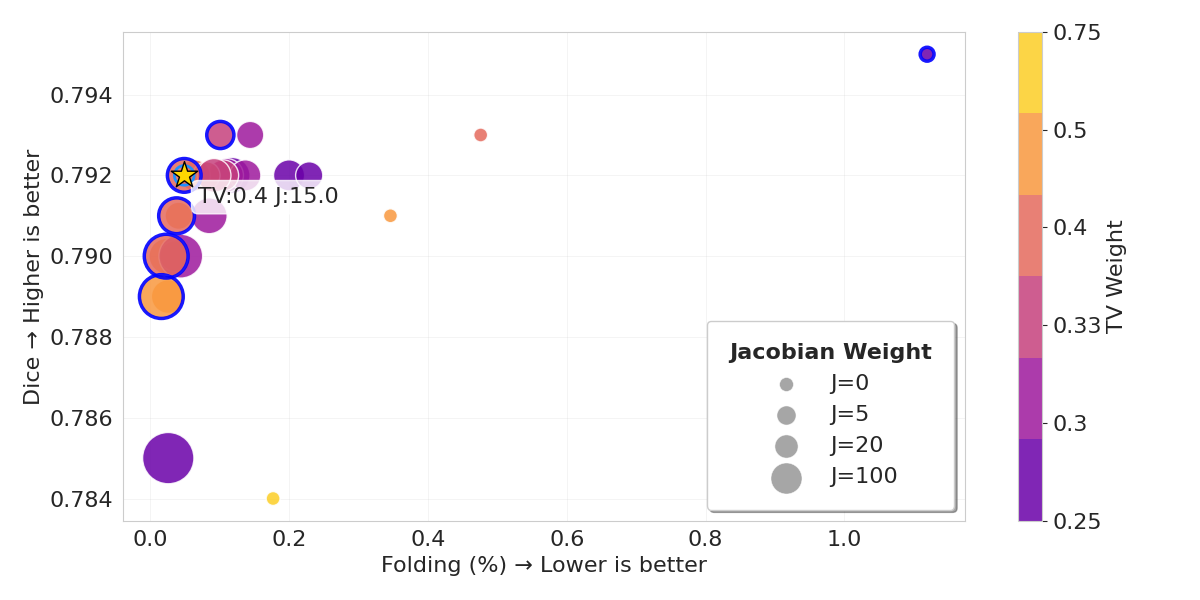}\vspace{-6pt}
    \caption{Ablation 
 of loss function weights for the RandKAN-IDIR model on the OASIS-1 dataset. Each point represents one configuration. Color encodes the TV weight $\lambda$, circle size encodes the Jacobian determinant regularization weight $\gamma$, blue outlines indicate Pareto-frontier configurations, and the star marks the selected final setting $\lambda = 0.4$, $\gamma = 15$.}
    \label{fig:loss_weights}
\end{figure}

 In this study, we emphasize the importance of Jacobian determinant regularization in our method. Figure~\ref{fig:loss_weights} illustrates the results for the RandKAN-IDIR model on the OASIS-1 dataset across 26 different loss function weight configurations. Circle size encodes the Jacobian determinant regularization weight $\gamma$, with the smallest circles corresponding to $\gamma = 0$, color encoding the TV weight $\lambda$. Combinations with $\gamma = 0$ exhibit poorer deformation regularity compared to others, even when increasing $\lambda$. Our selected final configuration, with $\lambda = 0.4$ and $\gamma = 15$, lies on the Pareto frontier, achieving a balance between Dice score and the amount of foldings.

\vspace{-3pt}

\subsubsection{Choice of $k$ and $K$ Parameters}
\label{sec:ablation_k_K}
\begin{figure}[H]
        \includegraphics[width=0.95\linewidth]{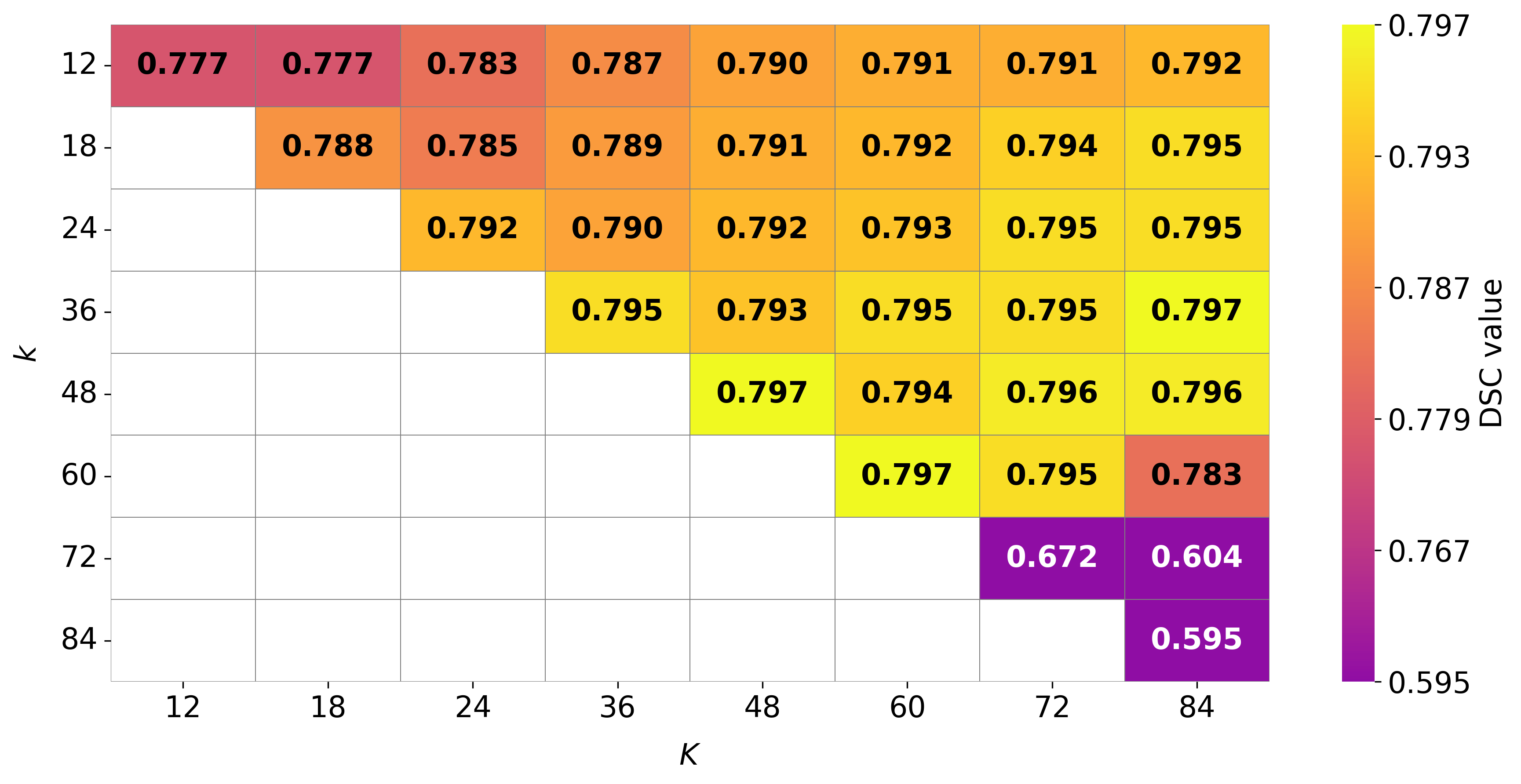}\vspace{-3pt}
    \caption{Ablation 
 study of $k$ and $K$ parameters for RandKAN-IDIR model on the OASIS-1 dataset.}
    \label{fig:k_K_ablation}
\end{figure}

Figure~\ref{fig:k_K_ablation} shows the results for the RandKAN-IDIR model on the OASIS-1 dataset with different values of \(k\) and \(K\). Overall, performance improves monotonically with increasing \(k\) up to 60, beyond which the model encounters convergence issues. For a fixed \(k\), models with randomized sampling (results for \(K > k\)) and sufficiently large value of \(K\) achieve better registration quality than models with sequential basis selection (results on the main diagonal with \(K = k\)). The improvement is especially notable for smaller values of \(k\). For example, the model with \(k = 12\) and \(K = 84\) outperforms the model with \(k = K = 12\) (equivalent to KAN-IDIR with \(D = 12\)) by 1.5\% DSC and matches the performance of the model with \(k = K = 24\). Moreover, our models can be scaled to reach slightly higher performance than reported in Table~\ref{tab:oasis_results}---up to 0.797 DSC---at the cost of slower runtime and increased VRAM consumption.

\subsubsection{Experiments with Different INR Architectures} To justify the choice of INR in our method, we conducted experiments with several popular INR models, including:

\begin{itemize}
    \item \textbf{ChebyKAN} \citep{sidharth2024chebyshev}—the architecture used in our KAN-IDIR model;
    \item \textbf{Cheby2KAN}—a ChebyKAN model using Chebyshev polynomials of the second kind as basis functions;
    \item \textbf{Efficient-KAN}—an efficient implementation of classical spline-based KANs \citep{liu2024kan};
    \item \textbf{FastKAN} \citep{li2024kolmogorovarnold}—a fast KAN variant that replaces third-order B-spline bases with radial basis functions (RBFs);
    \item \textbf{SIREN} \citep{sitzmann2020siren}—the architecture used in IDIR \citep{wolterink2022implicit};
    \item\item \textbf{SL\boldmath{$^{2}$A}} \citep{rezaeian2025sl2a}—a novel INR architecture that combines single-layer Chebyshev KAN and MLP with standard ReLU activations.
\end{itemize}

For all models, we tuned the learning rate schedule to achieve optimal performance. All KAN-based models share the same architecture: two hidden layers with 70 neurons each. For SIREN, we adopted the architecture used in IDIR, and for SL$^{2}$A we retained the original configuration. The number of basis functions ($D$ in ChebyKAN, \texttt{degree} in 
 SL$^{2}$A) and grid points (\texttt{grid} in Efficient-KAN and FastKAN) in KAN-based models was selected to approximately match the total parameter count across all networks. To ensure a fair comparison, we used the loss function from the main paper as the training objective for all methods. All models were trained with a batch size of 10{,}000 points until convergence.

The results of comparing the different INR architectures on the OASIS-1 dataset are shown in Table~\ref{table:kan_inr_ablation}. ChebyKAN---the backbone of our KAN-IDIR method---achieves the best performance in terms of both registration accuracy and runtime. Other INR baselines perform worse than ChebyKAN, which confirms the choice of this architecture for our method. Notably, the alternative KAN variants demonstrate considerably higher runtimes and VRAM usage than ChebyKAN, highlighting the computational efficiency of Chebyshev polynomials compared to other families of basis functions.

\begin{table}[H]
\caption{Evaluation of various INR models on the OASIS-1 dataset. Values in parentheses denote standard deviations. Results for the ChebyKAN architecture correspond to the KAN-IDIR model. The best scores are shown in \textbf{bold}, and the second best in \textit{italics}.}\vspace{-3pt}
\small
\label{table:kan_inr_ablation}
\centering
{%
\renewcommand{\arraystretch}{1.2}
\begin{tabularx}{\textwidth}{C|ccccc}
\noalign{\hrule height 1pt} 
\textbf{INR Architecture }                                                & \textbf{DSC ↑}                 & \textbf{HD95 ↓}               & \textbf{NJD, \% ↓}              & \textbf{VRAM, MB ↓} & \textbf{Runtime, s ↓} \\ \hline
ChebyKAN ($D=12$)                                     & 0.777 (0.17)          & 2.08 (0.33)          & \textbf{0.012 (0.01)} & 1412             & \textit{43.1}  \\
ChebyKAN ($D=28$)                                     & \textbf{0.793 (0.16)} & \textbf{1.96 (0.32)} & 0.048 (0.03)           & 2248             & 63.3           \\
Cheby2KAN ($D=28$)                                     & \textbf{0.793 (0.17)} & \textbf{1.96 (0.33)} & 0.124 (0.06)           & 4380             & 137.8           \\
Efficient-KAN ($\text{grid}=24$)                         & 0.768 (0.18)          & 2.16 (0.37)          & 0.029 (0.04)          & 4134             & 317.0          \\
Efficient-KAN ($\text{grid}=48$)                         & 0.787 (0.17)          & 2.00 (0.33)          & 0.198 (0.08)          & 7236             & 514.7          \\
FastKAN ($\text{grid}=24$)                                & 0.768 (0.18)          & 2.15 (0.35)          & \textit{0.024 (0.03)} & 2322             & 84.4           \\
FastKAN ($\text{grid}=48$)                               & 0.790 (0.17)          & \textit{1.99 (0.34)} & 0.301 (0.12)          & 3838             & 146.3          \\
SIREN                                                 & 0.784 (0.17)          & 2.02 (0.33)          & 0.040 (0.03)          & \textbf{792}     & \textbf{39.5}  \\
SL$^{2}$A ($\text{degree}=40$)  & \textit{0.791 (0.17)} & \textit{1.99 (0.33)} & 0.070 (0.05)           & \textit{1016}    & 72.1           \\
SL$^{2}$A ($\text{degree}=72$) & \textit{0.791 (0.17)} & 2.00 (0.33)          & 0.732 (0.44)          & 1076             & 74.0           \\ \noalign{\hrule height 1pt} 
\end{tabularx}
}
\end{table}

\section{Discussion}
\label{sec:discussion}

\subsection{Biomedical Relevance and Potential Applications}
Deformable image registration is a critical component of many clinically oriented image analysis pipelines, providing spatial correspondence that underlies meaningful interpretation of downstream measurements. For example, in lung CT, deformable registration enables the computation of ventilation maps that have been investigated for correlating regional lung function with radiation-induced toxicity and clinical outcomes \cite{vinogradskiy2013ventilation}. In cardiac MRI, registration‑based motion tracking has been shown to enable quantitative evaluation of myocardial motion and deformation across the full cardiac cycle, which supports assessment of regional cardiac function beyond simple global volumetric indices \cite{Qiao2020registration}. In brain MRI, longitudinal registration is a fundamental step for tracking regional structural changes over time and is routinely used in studies of neurodegenerative progression and brain atrophy \cite{reuter2012within}.

Our approach enhances the accuracy and plausibility of deformable image registration across multiple organs and imaging modalities, ensuring that downstream steps (from segmentation to functional or longitudinal analysis) operate on well-aligned, clinically meaningful data. Our models are pairwise-trained, allowing direct use in practical settings even when only a few images are available. In addition, RandKAN-IDIR demonstrates higher computational efficiency than other pairwise registration methods, making it suitable for scenarios with limited computational resources. The reduced seed-dependent variability makes single-run usage more reliable than prior INR baselines, which is important for reproducible clinically oriented research workflows, although clinical deployment would still require task-specific validation.
This combination of seed-stable optimization and modest compute/memory requirements makes per-pair INR-based DIR feasible for longitudinal and functional studies in data-scarce or resource-constrained research environments.

\subsection{Limitations}
\label{sec:limitations_ethics}
Our models optimize a deformation field per image pair, which can be slower at test time than a trained feed-forward regressor, especially when many pairs must be processed. As with other intensity-driven registration approaches, performance may degrade under strong appearance shifts (e.g., noise, artifacts, inconsistent preprocessing, or out-of-distribution protocols). Practical performance can also depend on optimization and regularization choices; transferring to new sites, scanners, cohorts, or tasks may require tuning and additional validation (e.g., uncommon pathology or very large deformations).

\subsection{Ethical Considerations}
Our goal is to support clinically oriented analysis pipelines; however, we do not report prospective clinical workflow studies, and decision-support use would require task-specific validation, monitoring, and human oversight. Medical imaging data should be used under appropriate governance (e.g., de-identification, consent where applicable, and dataset-specific approvals/agreements). Because benchmarks may not fully represent real-world populations and acquisition conditions, performance may vary across subgroups; evaluating robustness and potential bias is important before high-stakes use. Finally, to mitigate misuse and over-interpretation beyond validated settings, we encourage reporting failure modes/uncertainty and following regulatory and institutional requirements.

\section{Conclusions}

This paper presents KAN-IDIR and RandKAN-IDIR, a KAN-based INR framework targeting pairwise-optimized, resolution\-/independent DIR with an explicit focus on seed stability and resource efficiency.
By integrating randomized basis sampling into KAN-IDIR, RandKAN-IDIR reduces computational complexity while preserving near-equivalent registration quality. Across three medical imaging datasets covering different organs and modalities, both models provide the strongest overall performance among pairwise-optimized neural registration methods, while KAN-IDIR also achieves the highest DSC and best deformation regularity among all compared methods on ACDC.

Looking ahead, the proposed framework opens directions for further KAN architecture optimization and extension to broader medical image analysis tasks. Its combination of computational efficiency, initialization robustness, and instance-specific adaptability makes it potentially useful for research workflows and clinically oriented analysis pipelines after additional validation, especially when working with limited datasets. Prospective validation on multi-institutional data is still needed to assess generalizability across scanners and acquisition protocols.

\vspace{6pt}

\authorcontributions{Conceptualization, N.A.D., M.O.Z. and D.V.S.; methodology, N.A.D., M.O.Z. and D.V.S.; software, N.A.D. and M.O.Z.; validation, N.A.D., M.O.Z. and D.V.S.; formal analysis, N.A.D. and M.O.Z.; investigation, N.A.D., M.O.Z. and D.V.S.; writing---original draft preparation, N.A.D. and M.O.Z.; writing---review and editing, N.A.D., M.O.Z. and D.V.S.; visualization, N.A.D. and M.O.Z.; supervision, D.V.S.; project administration, D.V.S. All authors have read and agreed to the published version of the manuscript.}

\funding{This work was supported by the Ministry of Economic Development of the Russian Federation in accordance with the subsidy agreement (agreement identifier 000000C313925P4H0002; grant No. 139-15-2025-012).}

\institutionalreview{Ethical review and approval were not required for this study, as it used only publicly available, de-identified benchmark datasets.}

\informedconsent{Patient consent was not required for this study, as no new human subject data were collected and only publicly available, de-identified benchmark datasets were used.}

\dataavailability{Data Availability Statement: This study did not generate a new clinical or imaging dataset. The experiments were conducted using publicly available third-party benchmark datasets: DIR-Lab, OASIS-1, and ACDC, which are available from their original providers under their respective access conditions. These datasets are described, with references to the original sources, in the Datasets and Evaluation Metrics section of the manuscript. The source code for the proposed method is available at 
 \url{https://github.com/anac0der/KAN-IDIR} (accessed on 30th of June 2026).}

\acknowledgments{During the preparation of this manuscript, the authors used DeepSeek-R1-0528 
 (DeepSeek, \url{https://www.deepseek.com}) for language refinement and formatting. The authors have reviewed and edited all output and take full responsibility for the content of this publication.}

\conflictsofinterest{The authors declare no conflicts of interest.}

\abbreviations{Abbreviations}{
The following abbreviations are used in this manuscript:
\\

\noindent
\begin{tabular}{@{}ll}
DIR & Deformable image registration\\
INR & Implicit neural representation\\
KAN & Kolmogorov--Arnold network\\
CNN & Convolutional neural network\\
MLP & Multi-layer perceptron\\
CT & Computed tomography\\
MRI & Magnetic resonance imaging\\
TRE & Target registration error\\
DSC & Dice similarity coefficient\\
HD95 & 95th percentile Hausdorff distance\\
NJD & Negative Jacobian determinant\\
VRAM & Video random-access memory
\end{tabular}
}

\appendixtitles{yes}
\appendixstart
\appendix
\renewcommand{\theHequation}{\thesection.\arabic{equation}}
\section[\appendixname~\thesection]{Adaptive Learning of Basis Functions}\label{app1}
\vspace{-6pt}
\begin{table}[H]
\caption{Comparison 
 of A-KAN-IDIR and RandKAN-IDIR on DIR-Lab, OASIS-1, and ACDC datasets. The best result per metric is highlighted in \textbf{bold}.}\vspace{-3pt}
\label{akan_randkan}
\footnotesize
\centering
{%
\renewcommand{\arraystretch}{1.2}
\begin{tabularx}{\textwidth}{L|cc|ccc|ccc}
\noalign{\hrule height 1pt} 
\multirow{2.1}{*}{\textbf{Method}} & \multicolumn{2}{c|}{\textbf{DIR-Lab}} & \multicolumn{3}{c|}{\textbf{OASIS-1}} & \multicolumn{3}{c}{\textbf{ACDC}} \\
\cline{2-9}
 & \textbf{Avg. TRE} \boldmath{$\downarrow$} & \textbf{NJD} \boldmath{$\downarrow$} \textbf{(\%)} & \textbf{DSC} \boldmath{$\uparrow$} & \textbf{HD95} \boldmath{$\downarrow$} & \textbf{NJD} \boldmath{$\downarrow$} \textbf{(\%)} & \textbf{DSC} \boldmath{$\uparrow$} & \textbf{HD95} \boldmath{$\downarrow$} & \textbf{NJD} \boldmath{$\downarrow$} \textbf{(\%)} \\
\hline
A-KAN-IDIR & 1.02 & \textbf{0.006} & 0.786 & 2.01 & 0.135 & \textbf{0.811} & \textbf{7.76} & 8 $\times~10^{-5}$ \\
RandKAN-IDIR & \textbf{0.99} & \textbf{0.006} & \textbf{0.792} & \textbf{1.97} & \textbf{0.049 }& 0.810 & 7.88 & \textbf{4 \boldmath{$\times~10^{-5}$}} \\
\noalign{\hrule height 1pt} 
\end{tabularx}
}

\end{table}

\subsection{Description of A-KAN-IDIR vs. RandKAN-IDIR}
\label{sec:akanidir}

\textbf{A-KAN-IDIR architecture.}
Here we describe our A-KAN-IDIR model with adaptive learning of basis function indices.

We assign a learnable weight to each function in the basis function set, with each weight reflecting the importance of its corresponding basis function. The forward pass through the A-KAN-IDIR layer is formulated as follows:

\begin{equation} \label{eq_akan}
[\mathbf{y}(\mathbf{x})]_{b,o} = \sum_{i=1}^{N_{in}} \sum_{d \in \mathcal{D}_w} \text{sigmoid}(w_d) \cdot T_{d}(\mathbf{x}_{b,i} ) \cdot \mathbf{C}_{i,o,d},
\end{equation}
where $w_{d}$ represents the learnable weight, constrained to $(0, 1)$ with the sigmoid function, and $\mathcal{D}_w$ denotes the set of indices for basis functions selected at each iteration. To construct $\mathcal{D}_w$, we employ the Noisy Top-$k$ algorithm, which selects the indices corresponding to the top-$k$ values of the noisy weight vector:
\begin{equation} \label{noisy_topk}
\mathcal{D}_w = \text{TopkIndices}(\{w_i + n_i \mid w_i \in \mathbf{w}\}),
\end{equation}
where $\mathbf{w}$ denotes the vector of all weights, and $n_i$ represents Gaussian noise added to each weight $w_i$ to prevent convergence to local minima early in the training process. We model $n_i$ as Gaussian noise with a standard deviation that decreases from 0.3 at the start of training to zero by 75\% of the iterations. During the final 25\% of iterations, we freeze the learnable weights and train only the Chebyshev polynomial coefficients.

To achieve a richer parameterization of learnable basis function weights, we adopt the Deep Image Prior \citep{ulyanov2018deep} approach, where the target image is learned from a fixed noise vector. We learn the weight vector $\mathbf{w}$ for each layer from a unique fixed vector $\mathbf{z}$, sampled from a standard Gaussian distribution. The mapping from $\mathbf{z}$ to $\mathbf{w}$ is learned using a two-layer MLP. The values of $\mathbf{z}$, $\mathbf{w}$, and the MLP weights are layer-specific and not shared across~layers.

The overall forward pass pipeline for A-KAN-IDIR layer is as follows:
\begin{enumerate}
    \item Pass $\mathbf{z}$ to MLP and obtain the $\mathbf{w}$ values: $\mathbf{w} = \text{MLP}(\mathbf{z})$;
    \item Sample noise vector $\mathbf{n}$;
    \item Select set of indices $\mathcal{D}_w$ according to Equation~(\ref{noisy_topk});
    \item Constrain values of $\mathbf{w}$ with indices from $\mathcal{D}_w$ with sigmoid function;
    \item Perform forward pass according to Equation~(\ref{eq_akan}).
\end{enumerate}

\textbf{Comparison of A-KAN-IDIR vs RandKAN-IDIR.}
Table~\ref{akan_randkan} presents a comparative evaluation of A-KAN-IDIR and RandKAN-IDIR across the DIR-Lab, OASIS-1, and ACDC datasets, with metrics including Average TRE, DSC, HD95, and NJD. RandKAN-IDIR outperforms A-KAN-IDIR in most metrics across the datasets, achieving better registration quality and deformation regularity. However, on the ACDC dataset, A-KAN-IDIR achieves a slightly higher Dice score and better HD95. Additionally, A-KAN-IDIR exhibits approximately 20\% higher runtime (e.g., 51.2 s vs. 43.1 s on OASIS-1) and VRAM consumption (e.g., 2.64 Gb vs. 2.2 Gb on OASIS-1) compared to RandKAN-IDIR. These results suggest that A-KAN-IDIR’s adaptive basis selection mechanism introduces higher computational overhead compared to RandKAN-IDIR’s simpler randomized basis sampling, which provides superior registration quality and efficiency for deformable image registration in our experimental setup.

\section[\appendixname~\thesection]{Additional Experimental Results}

\subsection{Baselines Implementation Details and Parameters Selection}
\label{sec:baseline_impl}

For IDIR \citep{wolterink2022implicit}, ccIDIR \citep{vanharten2023robust}, NODEO \citep{wu2021nodeo}, and SINR \citep{sideri2024sinr} we extended the original implementations to all three datasets: OASIS-1, and ACDC for IDIR and ccIDIR; DIR-Lab, OASIS-1, and ACDC for SINR; DIR-Lab and ACDC for NODEO). We ensured that our modifications preserved performance and adhered strictly to the original implementations.

For SINR and NODEO, we identified and fixed several substantial bugs in both codebases that prevented execution on datasets beyond those in the original papers. Pending paper acceptance, we will submit these fixes as pull requests to the respective repositories.

IDIR, ccIDIR, and NODEO were executed using their default parameter configurations without modification. We conducted extensive hyperparameter tuning for these methods on the datasets not used in their original publications; however, this did not yield any substantial performance improvements. 

For SINR \citep{sideri2024sinr}, we revised the original stopping criterion (NJD $>$ 0.9\%) due to observed cases where NJD values initialized high and decreased during optimization. After testing iterations at 300, 500, and 1500, we standardized to 500 iterations, as no significant improvements were observed at higher counts. Notably, SINR required additional cropping to landmark min-max coordinates on DIR-Lab to accommodate 48GB GPU memory constraints—the only method necessitating such adaptation.

VoxelMorph \citep{balakrishnan2019tmi} and TransMorph \citep{chen2022transmorph} were trained without weak supervision with segmentation labels. We adjusted their train--test splits to use the last 49 pairs (versus the original 19) to enhance statistical robustness and maintain consistency with other experiments.

For pTV \citep{vishnevskiy2016isotropic} and CorrMLP \citep{meng2024correlation}, we directly reproduced the authors' reported values, as these represent current state-of-the-art implementations for their respective datasets.

\subsection{Calculation of HD95 Metric} \label{sec:hd95}
Let $S_{pred}$ denote the predicted segmentation mask and $S_{gt}$ the ground-truth segmentation mask.  
Let $\partial S_{pred}$ and $\partial S_{gt}$ denote the corresponding sets of boundary points.

Define the directed surface distance sets as
\begin{equation}
D(\partial S_{pred}, \partial S_{gt}) = 
\left\{ \min_{y \in \partial S_{gt}} \|x - y\| \;\middle|\; x \in \partial S_{pred} \right\},
\end{equation}
\begin{equation}
D(\partial S_{gt}, \partial S_{pred}) = 
\left\{ \min_{x \in \partial S_{pred}} \|y - x\| \;\middle|\; y \in \partial S_{gt} \right\}.
\end{equation}

The symmetric Hausdorff Distance (HD) is defined as the maximum distance in the union of the directed surface distance sets:
\begin{equation}
HD(S_{pred}, S_{gt}) = 
\max \Big( D(\partial S_{pred}, \partial S_{gt}) \;\cup\; D(\partial S_{gt}, \partial S_{pred}) \Big).
\end{equation}

To reduce sensitivity to outliers, the 95th percentile Hausdorff Distance (HD95) is defined as the 95th percentile of the union of the distance sets:
\begin{equation}
HD95(S_{pred}, S_{gt}) = 
P_{95} \Big( D(\partial S_{pred}, \partial S_{gt}) \;\cup\; D(\partial S_{gt}, \partial S_{pred}) \Big),
\end{equation}
where $P_{95}(\cdot)$ denotes the 95th percentile operator.

\subsection{Folding Severity Analysis}
\label{app:jacobian_severity}

Table~\ref{tab:oasis_folding_severity} reports additional folding-severity metrics for the OASIS-1 experiment. These metrics complement NJD by quantifying the magnitude of negative Jacobian determinants and the volume of severe negative-Jacobian regions.

\begin{table}[H]
\centering
\caption{Folding-severity analysis on the OASIS-1 dataset. Values are reported as mean (standard deviation). The best 
 result is highlighted in \textbf{bold}, and the second best in \textit{italics}. Proposed methods are marked with asterisk (*).}\vspace{-3pt}
\label{tab:oasis_folding_severity}
{%
\renewcommand{\arraystretch}{1.2}
\begin{tabularx}{\textwidth}{L|CCC}
\noalign{\hrule height 1pt} 
\textbf{Method} & \textbf{NJD, \%}  \boldmath{$\downarrow$} & \textbf{MeanNegJ}  \boldmath{$\downarrow$} & \textbf{SevereJ}\boldmath{$_{0.1}$}\textbf{, \%} \boldmath{$\downarrow$} \\
\hline
VoxelMorph & 0.530 (0.153) & 0.277 (0.031) & 0.333 (0.099) \\
TransMorph & 0.526 (0.142) & 0.279 (0.030) & 0.335 (0.089) \\
NODEO & \textbf{0.011 (0.008)} & \textbf{0.066 (0.021)} & \textbf{0.002 (0.002)} \\
SINR & 2.004 (0.521) & 0.359 (0.046) & 1.350 (0.362) \\
IDIR & \textit{0.041 (0.029)} & 0.096 (0.047) & \textit{0.015 (0.014)} \\
ccIDIR &  0.345 (0.117) & 0.301 (0.079) & 0.200 (0.074) \\
KAN-IDIR (28) * & 0.046 (0.027) & 0.101 (0.019) & 0.017 (0.012) \\
RandKAN-IDIR * & 0.043 (0.033) & \textit{0.095 (0.019)} &  \textit{0.015 (0.015)}\\
\noalign{\hrule height 1pt} 
\end{tabularx}
}
\end{table}

\subsection{Additional Quantitative Results}

\subsubsection{DIR-Lab Per-Case Results}
In Table \ref{dirlab_full_table} we show the full per-case results for DIR-Lab dataset, following the practice for this dataset in \citep{wolterink2022implicit,vanharten2023robust}.

\begin{table}[H]
\centering
\caption{Comparison 
 of proposed models with uniGradICON \cite{uniGradIcon} on OASIS and ACDC datasets. The best results are highlighted in \textbf{bold, and the second best in \textit{italics}}.}\vspace{-3pt}
\renewcommand{\arraystretch}{1.2}
\begin{tabularx}{\textwidth}{L|CC|CC}
\noalign{\hrule height 1pt} 
\multirow{2}{*}{\textbf{Model}} & \multicolumn{2}{c|}{\textbf{OASIS}} & \multicolumn{2}{c}{\textbf{ACDC}} \\
 & \textbf{DSC}  \boldmath{$\uparrow$} & \textbf{HD95}  \boldmath{$\downarrow$} & \textbf{DSC}  \boldmath{$\uparrow$} & \textbf{HD95}  \boldmath{$\downarrow$} \\
\hline
KAN-IDIR        & \textbf{0.793} & \textbf{1.96} & \textbf{0.814} & \textbf{7.39} \\
RandKAN-IDIR    & \textit{0.792} & \textit{1.97 }& \textit{0.810} &  \textit{7.88}\\
uniGradICON     & 0.791 & \textbf{1.96 }& 0.798 &  9.29 \\
\noalign{\hrule height 1pt} 
\end{tabularx}
\label{tab:unigrad_results}
\end{table}

\subsubsection{Comparison with Modern Foundational Models}
In Table~\ref{tab:unigrad_results}, we present a comparison of the proposed KAN-IDIR and RandKAN-IDIR models with uniGradICON~\cite{uniGradIcon}, a recent foundational model for deformable image registration.  Our models achieve performance comparable to uniGradICON on OASIS-1 and outperform it on ACDC, highlighting the effectiveness of the proposed approach.

\subsection{Additional Figures and Visualizations}

\begin{figure}[H]
\begin{adjustwidth}{-\extralength}{0cm}
\centering
\includegraphics[width=0.99\linewidth]{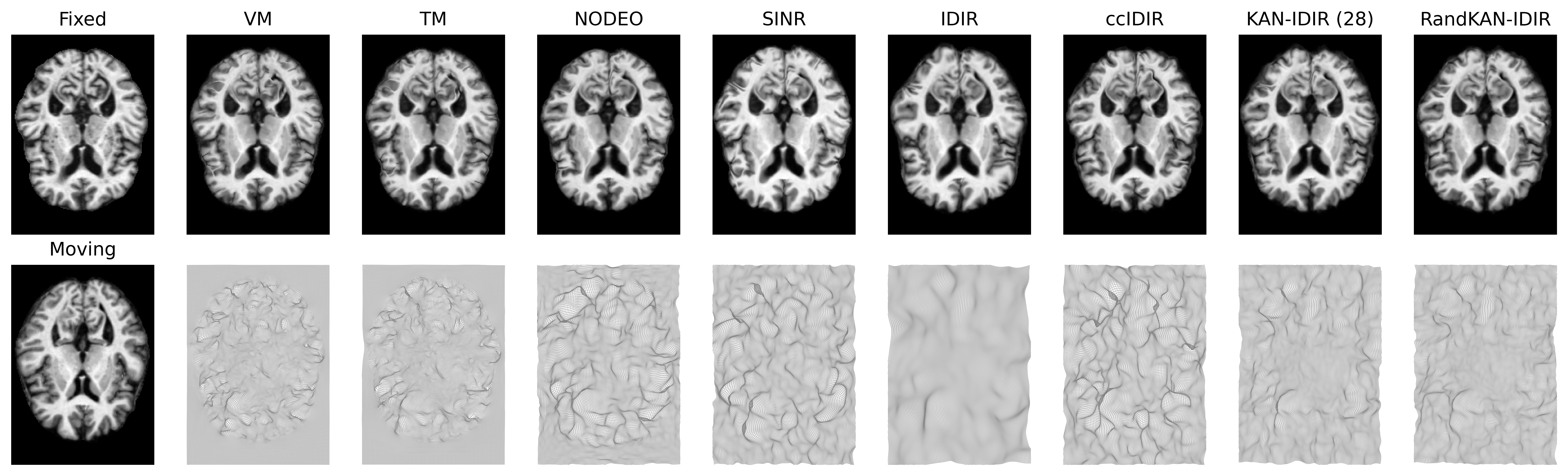}
\end{adjustwidth}
\caption{Visual comparison between different image registration models on OASIS-1 dataset. The first column consists of fixed image and moving image. The other columns consist (from top to bottom) of warped moving images obtained from the model and deformation grid.}
\label{fig:oasis_collage}
\end{figure}

\begin{figure}[H]
\begin{adjustwidth}{-\extralength}{0cm}
\centering
\includegraphics[width=0.99\linewidth]{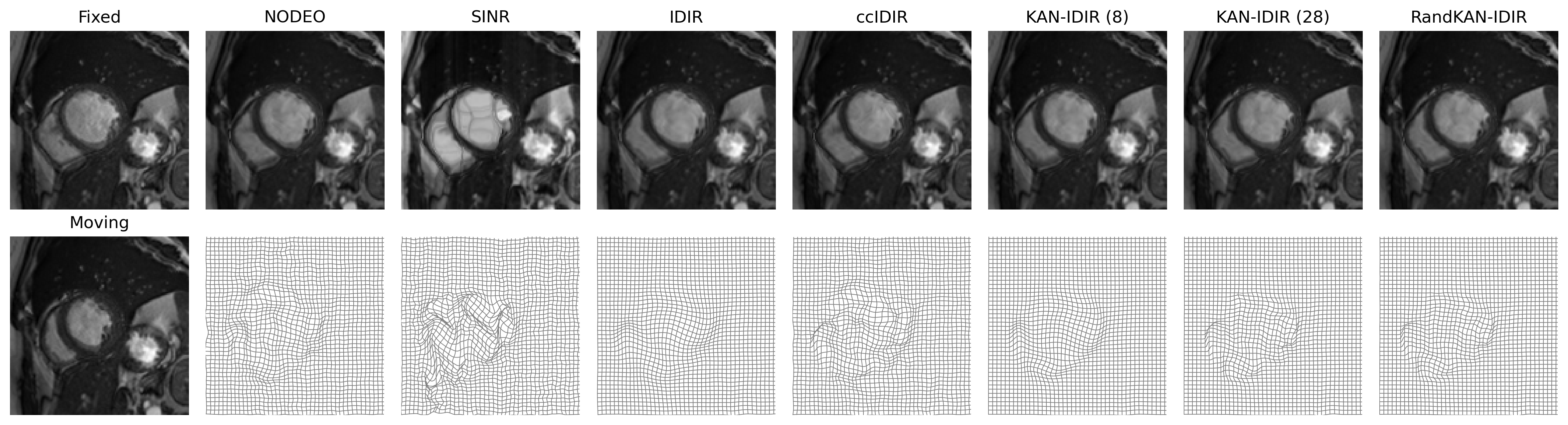}
\end{adjustwidth}
\caption{Visual comparison between different image registration models on ACDC dataset. The first column consists of fixed images and moving images. Other columns consist (from top to bottom) of warped moving image obtained from model and deformation grid.}
\label{fig:acdc_collage}
\end{figure}

\textbf{Visualization of registration results on brain and cardiac MRI.} 
Figures \ref{fig:oasis_collage} and \ref{fig:acdc_collage} present a qualitative comparison of different methods on the OASIS-1 and ACDC datasets. We extract a 2D slice from the 3D image and visualize the deformation by applying it to a corresponding 2D rectangular grid. The results demonstrate that our method produces smooth deformation fields, yielding images without visual artifacts.

\begin{figure}[H]
\begin{adjustwidth}{-\extralength}{0cm}
\centering
\includegraphics[width=0.9\linewidth]{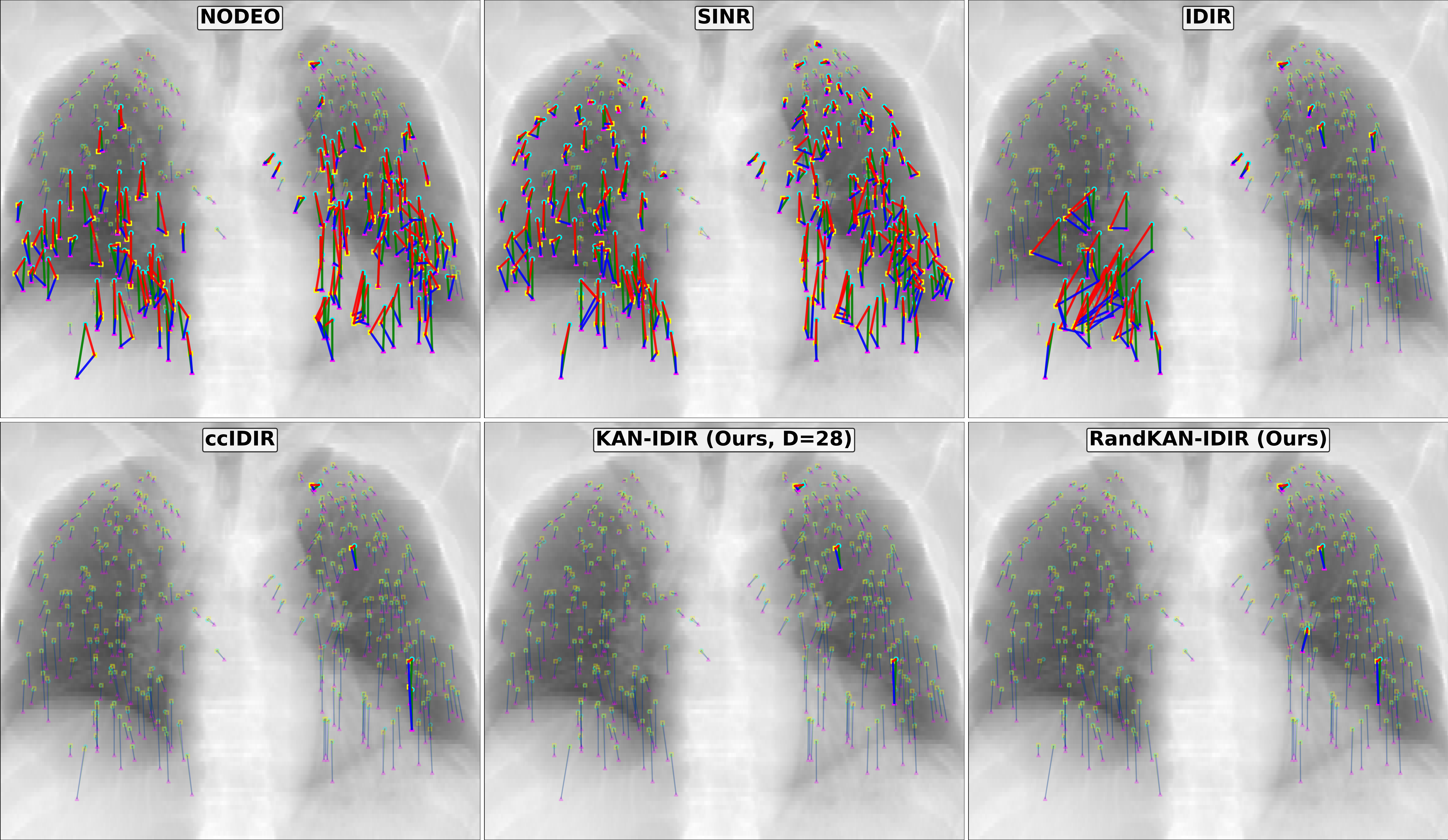}
\end{adjustwidth}
\caption{Landmark alignment in DIR-Lab Dataset (Case 7) for fixed, moving, and moved images. Displacements are shown as: blue (method result, moved-to-fixed), green (ground truth, moving-to-fixed), and red (error, moved-to-fixed). Landmarks with errors exceeding 3 mm are highlighted in bold, while others are displayed as transparent. Results reflect each method's suboptimal seed (maximum mean TRE across 10 runs). Proposed KAN-IDIR and RandKAN-IDIR still achieve the fewest outliers, while the other methods demonstrate significant increase. }
\label{dirlab_results_max_seed}
\end{figure}

\textbf{Visual Appearance of Registration Results for the Suboptimal Seed.}
The visual appearance of the registration results that confirms the best performance of the proposed models in terms of seed-dependent stability is demonstrated in Figure \ref{dirlab_results_max_seed}. The proposed RandKAN-IDIR and KAN-IDIR maintain consistent outlier counts across seeds, with results similar to the optimal seed in Figure \ref{dirlab_results}.

\vspace{-6pt}

\vspace{-6pt}

\begin{table}[H]
\caption{Evaluation 
 of TRE (lower better) 
 in mm for baselines and proposed methods on DIR-Lab dataset. Values in parentheses denote standard deviations across landmarks for each case and over all landmarks for the average row. The best result is highlighted in \textbf{bold}, and the second best in \textit{italics}.}
\footnotesize
\label{dirlab_full_table}
{%
\renewcommand{\arraystretch}{1.2}
\begin{tabularx}{\textwidth}{c|c|cccccc}
\noalign{\hrule height 1pt} 
\textbf{Case} & \textbf{pTV} & \textbf{NODEO} & \textbf{SINR} & \textbf{IDIR} & \textbf{ccIDIR} & \begin{tabular}[c]{@{}c@{}}\textbf{KAN-IDIR}\\ \textbf{(\textit{D 
} = 28, Ours)}\end{tabular} & \begin{tabular}[c]{@{}c@{}}\textbf{RandKAN-IDIR}\\ \textbf{(\textit{k} = 12, \textit{K} = 84, Ours)}\end{tabular} \\ \hline
4DCT 01 & 0.76 (0.90) & 0.88 (0.94) & 1.59 (1.33) & 0.76 (0.94) & 0.83 (0.94) & 0.77 (0.91) & 0.78 (0.92) \\
4DCT 02 & 0.77 (0.89) & 1.08 (1.62) & 1.81 (1.90) & 0.76 (0.94) & 0.78 (0.93) & 0.77 (0.93) & 0.76 (0.92) \\
4DCT 03 & 0.90 (1.05) & 1.38 (2.13) & 3.36 (3.10) & 0.94 (1.02) & 1.02 (1.10) & 0.89 (1.02) & 0.91 (1.03) \\
4DCT 04 & 1.24 (1.29) & 2.93 (3.64) & 3.82 (3.55) & 1.32 (1.27) & 1.37 (1.36) & 1.30 (1.28) & 1.29 (1.26) \\
4DCT 05 & 1.12 (1.44) & 3.30 (4.53) & 4.76 (4.12) & 1.23 (1.47) & 1.25 (1.51) & 1.13 (1.41) & 1.13 (1.41) \\
4DCT 06 & 0.85 (0.89) & 4.52 (5.55) & 4.58 (3.30) & 1.09 (1.03) & 1.06 (1.09) & 0.94 (1.01) & 0.94 (1.01) \\
4DCT 07 & 0.80 (1.28) & 7.00 (7.93) & 6.74 (6.41) & 1.12 (1.00) & 0.97 (0.98) & 0.93 (0.96) & 0.94 (0.97) \\
4DCT 08 & 1.34 (1.93) & 10.47 (10.34) & 12.10 (8.71) & 1.21 (1.29) & 1.13 (1.40) & 1.11 (1.26) & 1.13 (1.28) \\
4DCT 09 & 0.92 (0.94) & 4.06 (4.42) & 3.44 (2.51) & 1.22 (0.95) & 1.02 (0.93) & 1.02 (0.96) & 1.03 (0.96) \\
4DCT 10 & 0.82 (0.89) & 3.65 (5.31) & 4.10 (4.12) & 1.01 (1.05) & 0.96 (0.98) & 0.97 (0.99) & 0.99 (0.98) \\ \hline
Average & \textbf{0.95 (1.15)} & 3.93 (6.07) & 4.63 (5.26) & 1.07 (1.10) & 1.04 (1.12) & \textit{0.98 (1.10)} & 0.99 (1.10) \\ \noalign{\hrule height 1pt} 
\end{tabularx}
}
\end{table}

\vspace{-6pt}


\begin{adjustwidth}{-\extralength}{0cm}

\reftitle{References}

\begin{thebibliography}{999}

\bibitem[Rueckert et~al.(1999)Rueckert, Sonoda, Hayes, Hill, Leach, and
  Hawkes]{Rueckert1999}
Rueckert, D.; Sonoda, L.I.; Hayes, C.; Hill, D.L.G.; Leach, M.O.; Hawkes, D.J. Nonrigid registration using free-form deformations: Application to breast MR images. \emph{IEEE Trans. Med. Imaging} \textbf{1999}, \emph{18}, 712--721. \url{https://doi.org/10.1109/42.796284}.

\bibitem[Beg et~al.(2005)Beg, Miller, Trouv{\'e}, and Younes]{Beg2005LDDMM}
Beg, M.F.; Miller, M.I.; Trouvé, A.; Younes, L. Computing large deformation metric mappings via geodesic flows of diffeomorphisms. \emph{Int. J. Comput. Vis.} \textbf{2005}, \emph{61}, 139--157. \url{https://doi.org/10.1023/B:VISI.0000043755.93987.aa}.

\bibitem[Thirion(1998)]{Thirion1998}
Thirion, J.P. Image matching as a diffusion process: An analogy with Maxwell’s demons. \emph{Med. Image Anal.} \textbf{1998}, \emph{2}, 243--260. \url{https://doi.org/10.1016/S1361-8415(98)80022-4}.

\bibitem[Klein et~al.(2009)Klein, Staring, Murphy, Viergever, and
  Pluim]{Klein2009Elastix}
Klein, S.; Staring, M.; Murphy, K.; Viergever, M.A.; Pluim, J.P.W. elastix: A Toolbox for Intensity-Based Medical Image Registration. \emph{IEEE Trans. Med. Imaging} \textbf{2009}, \emph{29}, 196--205. \url{https://doi.org/10.1109/TMI.2009.2035616}.

\bibitem[Avants et~al.(2011)Avants, Tustison, Song, Cook, Klein, and
  Gee]{Avants2011ANTs}
Avants, B.B.; Tustison, N.J.; Song, G.; Cook, P.A.; Klein, A.; Gee, J.C. A Reproducible Evaluation of ANTs Similarity Metric Performance in Brain Image Registration. \emph{NeuroImage} \textbf{2011}, \emph{54}, 2033--2044. \url{https://doi.org/10.1016/j.neuroimage.2010.09.025}.

\bibitem[Vishnevskiy et~al.(2016)Vishnevskiy, Gass, Szekely, Tanner, and
  Goksel]{vishnevskiy2016isotropic}
Vishnevskiy, V.; Gass, T.; Szekely, G.; Tanner, C.; Goksel, O. Isotropic total variation regularization of displacements in parametric image registration. \emph{IEEE Trans. Med. Imaging} \textbf{2016}, \emph{36}, 385--395. 
 \url{https://doi.org/10.1109/tmi.2016.2610583}.

\bibitem[Zhang and Fletcher(2019)]{Zhang2019FourierFLASH}
Zhang, M.; Fletcher, P.T. Fast diffeomorphic image registration via Fourier-approximated Lie algebras. \emph{Int. J. Comput. Vis.} \textbf{2019}, \emph{127}, 61--73. \url{https://doi.org/10.1007/s11263-018-1099-x}.

\bibitem[Balakrishnan et~al.(2019)Balakrishnan, Zhao, Sabuncu, Guttag, and
  Dalca]{balakrishnan2019tmi}
Balakrishnan, G.; Zhao, A.; Sabuncu, M.R.; Guttag, J.; Dalca, A.V. VoxelMorph: A Learning Framework for Deformable Medical Image Registration. \emph{IEEE Trans. Med. Imaging} \textbf{2019}, \emph{38}, 1788--1800. \url{https://doi.org/10.1109/TMI.2019.2897538}.

\bibitem[Chen et~al.(2022)Chen, Frey, He, Segars, Li, and
  Du]{chen2022transmorph}
Chen, J.; Frey, E.C.; He, Y.; Segars, W.P.; Li, Y.; Du, Y. TransMorph: Transformer for Unsupervised Medical Image Registration. \emph{Med. Image Anal.} \textbf{2022}, \emph{82}, 102615. \url{https://doi.org/10.1016/j.media.2022.102615}.

\bibitem[Mildenhall et~al.(2020)Mildenhall, Srinivasan, Tancik, Barron,
  Ramamoorthi, and Ng]{mildenhall2020nerf}
Mildenhall, B.; Srinivasan, P.P.; Tancik, M.; Barron, J.T.; Ramamoorthi, R.; Ng, R. NeRF: Representing Scenes as Neural Radiance Fields for View Synthesis. In \emph{Proceedings of the Computer Vision—ECCV 2020}; Lecture Notes in Computer Science; Springer: Berlin/Heidelberg, Germany, 
 2020; Volume 12346, pp. 405--421. \url{https://doi.org/10.1007/978-3-030-58452-8_24}.

\bibitem[Wolterink et~al.(2022)Wolterink, Zwienenberg, and
  Brune]{wolterink2022implicit}
Wolterink, J.M.; Zwienenberg, J.C.; Brune, C. Implicit Neural Representations for Deformable Image Registration. In Proceedings of the Medical Imaging with Deep Learning (MIDL), PMLR (Proceedings of Machine Learning Research), Zurich, Switzerland, 6-8 July 2022; Volume 172, pp. 1349--1359.

\bibitem[van Harten et~al.(2024{\natexlab{a}})van Harten, Stoker, and
  I{\v{s}}gum]{vanharten2023robust}
van Harten, L.D.; Stoker, J.; Išgum, I. Robust Deformable Image Registration Using Cycle-Consistent Implicit Representations. \emph{IEEE Trans. Med. Imaging} \textbf{2024}, \emph{43}, 784--793. \url{https://doi.org/10.1109/TMI.2023.3321425}.

\bibitem[van Harten et~al.(2024{\natexlab{b}})van Harten, Herten, Stoker, and
  Isgum]{vanharten2024deformable}
van Harten, L.; Herten, R.L.M.V.; Stoker, J.; Isgum, I. Deformable Image Registration with Geometry-Informed Implicit Neural Representations. In Proceedings of the Medical Imaging with Deep Learning (MIDL), PMLR (Proceedings of Machine Learning Research), Vienna, Austria, 21--27 July 2024; Volume 227, pp. 730--742.

\bibitem[Sitzmann et~al.(2020)Sitzmann, Martel, Bergman, Lindell, and
  Wetzstein]{sitzmann2020siren}
Sitzmann, V.; Martel, J.N.P.; Bergman, A.W.; Lindell, D.B.; Wetzstein, G. Implicit Neural Representations with Periodic Activation Functions. In Proceedings of the Conference on Neural Information Processing Systems (NeurIPS), Virtual, 6--12 December  2020; pp. 7461--7472.

\bibitem[Saragadam et~al.(2023)Saragadam, LeJeune, Tan, Balakrishnan,
  Veeraraghavan, and Baraniuk]{saragadam2023wire}
Saragadam, V.; LeJeune, D.; Tan, J.; Balakrishnan, G.; Veeraraghavan, A.; Baraniuk, R.G. WIRE: Wavelet Implicit Neural Representations. In Proceedings of the IEEE/CVF Conference on Computer Vision and Pattern Recognition (CVPR), Vancouver, BC, Canada, 18--22 June 2023; pp. 18507--18516.

\bibitem[Liu et~al.(2024)Liu, Zhu, Zhang, Fu, Deng, Ma, Guo, and
  Cao]{Liu_2024_CVPR}
Liu, Z.; Zhu, H.; Zhang, Q.; Fu, J.; Deng, W.; Ma, Z.; Guo, Y.; Cao, X. FINER: Flexible Spectral-Bias Tuning in Implicit Neural Representation by Variable-Periodic Activation Functions. In Proceedings of the IEEE/CVF Conference on Computer Vision and Pattern Recognition (CVPR), Seattle, WA, USA, 16--22 June 2024; pp. 2713--2722.

\bibitem[Liu et~al.(2025)Liu, Wang, Vaidya, Ruehle, Halverson,
  Solja{\v{c}}i{\'c}, Hou, and Tegmark]{liu2024kan}
Liu, Z.; Wang, Y.; Vaidya, S.; Ruehle, F.; Halverson, J.; Soljačić, M.; Hou, T.Y.; Tegmark, M. KAN: Kolmogorov-Arnold Networks. In Proceedings of the International Conference on Learning Representations (ICLR), Singapore, 24--28 April  2025.

\bibitem[Mehrabian et~al.(2024)Mehrabian, Mojarad~Adi, Heidari, and
  Hacihaliloglu]{mehrabian2024implicit}
Mehrabian, A.; Mojarad Adi, P.; Heidari, M.; Hacihaliloglu, I. Implicit Neural Representations with Fourier Kolmogorov-Arnold Networks. \emph{arXiv} \textbf{2024}, arXiv:cs.CV/2409.09323. \url{https://doi.org/10.48550/arXiv.2409.09323}.

\bibitem[Li et~al.(2025)Li, Zhang, Wang, Zhang, Li, and
  Shen]{li2025representing}
Li, L.; Zhang, L.; Wang, Z.; Zhang, F.; Li, Z.; Shen, Y. Representing Sounds as Neural Amplitude Fields: A Benchmark of Coordinate-MLPs and a Fourier Kolmogorov-Arnold Framework. In Proceedings of the AAAI Conference on Artificial Intelligence, Philadelphia, PA, USA, 27 February--2 March 2025; Volume 39, pp. 24458--24466. \url{https://doi.org/10.1609/aaai.v39i23.34624}.

\bibitem[Ramasinghe and Lucey(2022)]{ramasinghe2022beyond}
Ramasinghe, S.; Lucey, S. Beyond Periodicity: Towards a Unifying Framework for Activations in Coordinate-MLPs. In \emph{Proceedings of the Computer Vision—ECCV 2022}; Springer: Berlin/Heidelberg, Germany, 2022; pp. 142--158.

\bibitem[Vonderfecht and Liu(2024)]{vonderfecht2024predicting}
Vonderfecht, J.; Liu, F. Predicting the Encoding Error of SIRENs. \emph{arXiv} \textbf{2024}, arXiv:cs.CV/2410.21645. \url{https://doi.org/10.48550/arXiv.2410.21645}.

\bibitem[Goyal et~al.(2019)Goyal, Goyal, and Lall]{goyal2019learning}
Goyal, M.; Goyal, R.; Lall, B. Learning Activation Functions: A New Paradigm for Understanding Neural Networks. \emph{arXiv} \textbf{2019}, arXiv:cs.LG/1906.09529. \url{https://doi.org/10.48550/arXiv.1906.09529}.

\bibitem[Maxwell(1871)]{Maxwell1871}
Maxwell, J.C. \emph{Theory of Heat}; Longmans, Green and Co.: London, UK, 1871.

\bibitem[Dalca et~al.(2019)Dalca, Balakrishnan, Guttag, and
  Sabuncu]{dalca2019unsupervised}
Dalca, A.V.; Balakrishnan, G.; Guttag, J.; Sabuncu, M.R. Unsupervised learning of probabilistic diffeomorphic registration for images and surfaces. \emph{Med. Image Anal.} \textbf{2019}, \emph{57}, 226--236. \url{https://doi.org/10.1016/j.media.2019.07.006}.

\bibitem[Mok and Chung(2020{\natexlab{a}})]{mok2020large}
Mok, T.C.W.; Chung, A.C.S. Large deformation diffeomorphic image registration with laplacian pyramid networks. In \emph{Proceedings of the Medical Image Computing and Computer Assisted Intervention—MICCAI 2020}; Lecture Notes in Computer Science; Springer: Berlin/Heidelberg, Germany, 2020; Volume 12263, pp. 211--221. \url{https://doi.org/10.1007/978-3-030-59716-0_21}.

\bibitem[Mok and Chung(2020{\natexlab{b}})]{mok2020fast}
Mok, T.C.; Chung, A.C. Fast Symmetric Diffeomorphic Image Registration with Convolutional Neural Networks. In Proceedings of the IEEE/CVF Conference on Computer Vision and Pattern Recognition (CVPR), Seattle, WA, USA, 13--19 June 2020; \mbox{pp.~4644--4653}. \url{https://doi.org/10.1109/CVPR42600.2020.00470}.

\bibitem[Greer et~al.(2021)Greer, Kwitt, Vialard, and
  Niethammer]{greer2021icon}
Greer, H.; Kwitt, R.; Vialard, F.X.; Niethammer, M. ICON: Learning Regular Maps Through Inverse Consistency. In Proceedings of the IEEE/CVF International Conference on Computer Vision (ICCV), Montreal, BC, Canada, 11--17 October 2021; pp. 3396--3405.

\bibitem[Tian et~al.(2023)Tian, Greer, Vialard, Kwitt, Est{\'e}par, Rushmore,
  Makris, Bouix, and Niethammer]{tian2023gradicon}
Tian, L.; Greer, H.; Vialard, F.X.; Kwitt, R.; Estépar, R.S.J.; Rushmore, R.J.; Makris, N.; Bouix, S.; Niethammer, M. GradICON: Approximate Diffeomorphisms via Gradient Inverse Consistency. In Proceedings of the IEEE/CVF Conference on Computer Vision and Pattern Recognition (CVPR), Vancouver, BC, Canada, 18--22 June 2023; pp. 18084--18094. \url{https://doi.org/10.1109/CVPR52729.2023.01734}.

\bibitem[Pasenko and Davydov(2026)]{pasenko2026ctcf}
Pasenko, D.; Davydov, R. CTCF: A Three-Level Coarse-to-Fine Cascade for Unsupervised Deformable Medical Image Registration. \emph{Mach. Learn. Knowl. Extr.} \textbf{2026}, \emph{8}, 122. \url{https://doi.org/10.3390/make8050122}.

\bibitem[Meng et~al.(2024)Meng, Feng, Bi, and Kim]{meng2024correlation}
Meng, M.; Feng, D.; Bi, L.; Kim, J. Correlation-Aware Coarse-to-Fine MLPs for Deformable Medical Image Registration. In Proceedings of the IEEE/CVF Conference on Computer Vision and Pattern Recognition (CVPR), Seattle, WA, USA, 16--22 June 2024; pp. 9645--9654.

\bibitem[Jena et~al.(2024)Jena, Sethi, Chaudhari, and Gee]{jena2024deep}
Jena, R.; Sethi, D.; Chaudhari, P.; Gee, J. Deep Learning in Medical Image Registration: Magic or Mirage? \emph{Adv. Neural Inf. Process. Syst.} \textbf{2024}, \emph{37}, 108331--108353.

\bibitem[Tian et~al.(2024)Tian, Greer, Kwitt, Vialard, {San Jos{\'{e}}
  Est{\'{e}}par}, Bouix, Rushmore, and Niethammer]{uniGradIcon}
Tian, L.; Greer, H.; Kwitt, R.; Vialard, F.X.; San José Estépar, R.; Bouix, S.; Rushmore, R.; Niethammer, M. uniGradICON: A Foundation Model for Medical Image Registration. In \emph{Proceedings of the Medical Image Computing and Computer Assisted Intervention—MICCAI 2024}; Linguraru, M.G., Dou, Q., Feragen, A., Giannarou, S., Glocker, B., Lekadir, K., Schnabel, J.A., Eds.; Lecture Notes in Computer Science; Springer: Cham, Switzerland, 2024; Volume 15002, pp. 749--760. \url{https://doi.org/10.1007/978-3-031-72069-7_70}.

\bibitem[Castillo et~al.(2009)Castillo, Castillo, Guerra, Johnson, McPhail,
  Garg, and Guerrero]{castillo2009framework}
Castillo, R.; Castillo, E.; Guerra, R.; Johnson, V.E.; McPhail, T.; Garg, A.K.; Guerrero, T. A framework for evaluation of deformable image registration spatial accuracy using large landmark point sets. \emph{Phys. Med. Biol.} \textbf{2009}, \emph{54}, 1849. \url{https://doi.org/10.1088/0031-9155/54/7/001}.

\bibitem[Drozdov and Sorokin(2024)]{drozdov2024fnoreg}
Drozdov, N.A.; Sorokin, D.V. FNOReg: Resolution-Robust Medical Image Registration Method Based on Fourier Neural Operator. In \emph{Proceedings of the Pattern Recognition (ICPR 2024)}; Lecture Notes in Computer Science; Springer: Berlin/Heidelberg, Germany, 2024; Volume 15313, pp. 163--177. \url{https://doi.org/10.1007/978-3-031-78201-5_11}.

\bibitem[Byra et~al.(2023)Byra, Poon, Rachmadi, Schlachter, and
  Skibbe]{byra2023exploring}
Byra, M.; Poon, C.; Rachmadi, M.F.; Schlachter, M.; Skibbe, H. Exploring the performance of implicit neural representations for brain image registration. \emph{Sci. Rep.} \textbf{2023}, \emph{13}, 17334. 
 \url{https://doi.org/10.1038/s41598-023-44517-5}.

\bibitem[Wu et~al.(2022)Wu, Jiahao, Wang, Yushkevich, Hsieh, and
  Gee]{wu2021nodeo}
Wu, Y.; Jiahao, T.Z.; Wang, J.; Yushkevich, P.A.; Hsieh, M.A.; Gee, J.C. NODEO: A Neural Ordinary Differential Equation Based Optimization Framework for Deformable Image Registration. In Proceedings of the IEEE/CVF Conference on Computer Vision and Pattern Recognition (CVPR), New Orleans, LA, USA, 21--24 June 2022; pp. 20804--20813. 


\bibitem[Sideri-Lampretsa et~al.(2024)Sideri-Lampretsa, McGinnis, Qiu,
  Paschali, Simson, and Rueckert]{sideri2024sinr}
Sideri-Lampretsa, V.; McGinnis, J.; Qiu, H.; Paschali, M.; Simson, W.; Rueckert, D. SINR: Spline-Enhanced Implicit Neural Representation for Multi-Modal Registration. In Proceedings of the Medical Imaging with Deep Learning (MIDL), PMLR (Proceedings of Machine Learning Research), Vienna, Austria, 21--27 July 2024; Volume 250, pp. 1462--1474.

\bibitem[Cybenko(1989)]{cybenko1989approximation}
Cybenko, G. Approximation by superpositions of a sigmoidal function. \emph{Math. Control Signals Syst.} \textbf{1989}, \emph{2}, 303--314.

\bibitem[Hornik et~al.(1989)Hornik, Stinchcombe, and
  White]{hornik1989multilayer}
Hornik, K.; Stinchcombe, M.; White, H. Multilayer feedforward networks are universal approximators. \emph{Neural Netw.} \textbf{1989}, \emph{2}, 359--366.

\bibitem[Kolmogorov(1956)]{kolmogorov1957representation}
Kolmogorov, A.N. On the representation of continuous functions of several variables by superpositions of continuous functions of a smaller number of variables. \emph{Dokl. Akad. Nauk SSSR} \textbf{1956}, \emph{108}, 179--182.

\bibitem[Arnold(1957)]{arnold1957functions}
Arnold, V.I. On functions of three variables. \emph{Dokl. Akad. Nauk SSSR} \textbf{1957}, \emph{114}, 953--956.

\bibitem[Liu et~al.(2024)Liu, Ma, Wang, Matusik, and Tegmark]{Liu2024KAN2}
Liu, Z.; Ma, P.; Wang, Y.; Matusik, W.; Tegmark, M. KAN 2.0: Kolmogorov-Arnold Networks Meet Science. \emph{arXiv} \textbf{2024}, arXiv:cs.LG/2408.10205. \url{https://doi.org/10.48550/arXiv.2408.10205}.

\bibitem[Faroughi and Mostajeran(2025)]{faroughi2025neural}
Faroughi, S.A.; Mostajeran, F. Neural Tangent Kernel Analysis to Probe Convergence in Physics-Informed Neural Solvers: PIKANs vs. PINNs. \emph{arXiv} \textbf{2025}, arXiv:cs.LG/2506.07958. \url{https://doi.org/10.48550/arXiv.2506.07958}.

\bibitem[Sidharth et~al.(2024)Sidharth, Keerthana, Gokul, and
  Anas]{sidharth2024chebyshev}
Sidharth, S.S.; Keerthana, A.R.; Gokul, R.; Anas, K.P. Chebyshev Polynomial-Based Kolmogorov-Arnold Networks: An Efficient Architecture for Nonlinear Function Approximation. \emph{arXiv} \textbf{2024}, arXiv:cs.LG/2405.07200. \url{https://doi.org/10.48550/arXiv.2405.07200}.

\bibitem[Zhang et~al.(2015)Zhang, Xu, Yang, Li, and Zhang]{zhang2015survey}
Zhang, Z.; Xu, Y.; Yang, J.; Li, X.; Zhang, D. A survey of sparse representation: Algorithms and applications. \emph{IEEE Access} \textbf{2015}, \emph{3}, 490--530. \url{https://doi.org/10.1109/ACCESS.2015.2430359}.

\bibitem[Mallat and Zhang(1993)]{mallat1993matching}
Mallat, S.G.; Zhang, Z. Matching Pursuits with Time-Frequency Dictionaries. \emph{IEEE Trans. Signal Process.} \textbf{1993}, \emph{41}, 3397--3415. \url{https://doi.org/10.1109/78.258082}.

\bibitem[Aharon et~al.(2006)Aharon, Elad, and Bruckstein]{aharon2006ksvd}
Aharon, M.; Elad, M.; Bruckstein, A. K-SVD: An Algorithm for Designing Overcomplete Dictionaries for Sparse Representation. \emph{IEEE Trans. Signal Process.} \textbf{2006}, \emph{54}, 4311--4322. \url{https://doi.org/10.1109/TSP.2006.881199}.

\bibitem[Rudin et~al.(1992)Rudin, Osher, and Fatemi]{Rudin1992NonlinearTV}
Rudin, L.I.; Osher, S.; Fatemi, E. Nonlinear total variation based noise removal algorithms. \emph{Phys. D Nonlinear Phenom.} \textbf{1992}, \emph{60}, 259--268. \url{https://doi.org/10.1016/0167-2789(92)90242-F}.

\bibitem[Marcus et~al.(2007)Marcus, Wang, Parker, Csernansky, Morris, and
  Buckner]{marcus2007open}
Marcus, D.S.; Wang, T.H.; Parker, J.; Csernansky, J.G.; Morris, J.C.; Buckner, R.L. Open Access Series of Imaging Studies (OASIS): Cross-Sectional MRI Data in Young, Middle Aged, Nondemented, and Demented Older Adults. \emph{J. Cogn. Neurosci.} \textbf{2007}, \emph{19}, 1498--1507. \url{https://doi.org/10.1162/jocn.2007.19.9.1498}.

\bibitem[Hoopes et~al.(2022)Hoopes, Hoffmann, Greve, Fischl, Guttag, and
  Dalca]{hoopes2022learning}
Hoopes, A.; Hoffmann, M.; Greve, D.N.; Fischl, B.; Guttag, J.; Dalca, A.V. Learning the Effect of Registration Hyperparameters with HyperMorph. \emph{J. Mach. Learn. Biomed. Imaging} \textbf{2022}, \emph{1}, 003.

\bibitem[Hering et~al.(2023)Hering, Hansen, Mok, Chung, Siebert, H{\"a}ger,
  Lange, Kuckertz, Heldmann, Shao, Vesal, Rusu, Sonn, Estienne, Vakalopoulou,
  Han, Huang, Yap, Brudfors, Balbastre, Joutard, Modat, Lifshitz, Raviv, Lv,
  Jaouen, Visvikis, Fourcade, Rubeaux, Pan, Xu, Jian, De~Benetti, Wodzinski,
  Gunnarsson, Sj{\"o}lund, Grzech, Qiu, Li, Thorley, Duan, Gro{\ss}br{\"o}hmer,
  Hoopes, Reinertsen, Xiao, Landman, Huo, Murphy, Lessmann, Van~Ginneken,
  Dalca, and Heinrich]{hering2022learn2reg}
Hering, A.; Hansen, L.; Mok, T.C.W.; Chung, A.C.S.; Siebert, H.; Häger, S.; Lange, A.; Kuckertz, S.; Heldmann, S.; Shao, W.; et al. Learn2Reg: Comprehensive Multi-Task Medical Image Registration Challenge, Dataset and Evaluation in the Era of Deep Learning. \emph{IEEE Trans. Med. Imaging} \textbf{2023}, \emph{42}, 697--712. \url{https://doi.org/10.1109/TMI.2022.3213983}.

\bibitem[Bernard et~al.(2018)Bernard, Lalande, Zotti, Cervenansky, Yang, Heng,
  Cetin, Lekadir, Camara, Gonzalez~Ballester, Sanroma, Napel, Petersen,
  Tziritas, Grinias, Khened, Kollerathu, Krishnamurthi, Rohe, Pennec,
  Sermesant, Isensee, Jager, Maier-Hein, Full, Wolf, Engelhardt, Baumgartner,
  Koch, Wolterink, Isgum, Jang, Hong, Patravali, Jain, Humbert, and
  Jodoin]{bernard2018deep}
Bernard, O.; Lalande, A.; Zotti, C.; Cervenansky, F.; Yang, X.; Heng, P.A.; Cetin, I.; Lekadir, K.; Camara, O.; Gonzalez Ballester, M.A.; et al. Deep Learning Techniques for Automatic MRI Cardiac Multi-Structures Segmentation and Diagnosis: Is the Problem Solved? \emph{IEEE Trans. Med. Imaging} \textbf{2018}, \emph{37}, 2514--2525. \url{https://doi.org/10.1109/TMI.2018.2837502}.

\bibitem[Paszke et~al.(2019)Paszke, Gross, Massa, Lerer, Bradbury, Chanan,
  Killeen, Lin, Gimelshein, Antiga, Desmaison, Kopf, Yang, DeVito, Raison,
  Tejani, Chilamkurthy, Steiner, Fang, Bai, and Chintala]{paszke2019pytorch}
Paszke, A.; Gross, S.; Massa, F.; Lerer, A.; Bradbury, J.; Chanan, G.; Killeen, T.; Lin, Z.; Gimelshein, N.; Antiga, L.; et al. PyTorch: An Imperative Style, High-Performance Deep Learning Library. \emph{Adv. Neural Inf. Process. Syst.} \textbf{2019}, \emph{32}, 8024--8035.

\bibitem[Kingma and Ba(2015)]{kingma2014adam}
Kingma, D.P.; Ba, J. Adam: A Method for Stochastic Optimization. In Proceedings of the International Conference on Learning Representations (ICLR), San Diego, CA, USA, 7--9 May 2015.

\bibitem[Hofmanninger et~al.(2020)Hofmanninger, Prayer, Pan, Röhrich, Prosch,
  and Langs]{Hofmanninger2020AutomaticLungSegmentation}
Hofmanninger, J.; Prayer, F.; Pan, J.; Röhrich, S.; Prosch, H.; Langs, G. Automatic lung segmentation in routine imaging is primarily a data diversity problem, not a methodology problem. \emph{Eur. Radiol. Exp.} \textbf{2020}, \emph{4}, 50. 
 \url{https://doi.org/10.1186/s41747-020-00173-2}.

\bibitem[Li(2024)]{li2024kolmogorovarnold}
Li, Z. Kolmogorov-Arnold Networks Are Radial Basis Function Networks. \emph{arXiv} \textbf{2024}, arXiv:cs.LG/2405.06721. \url{https://doi.org/10.48550/arXiv.2405.06721}.

\bibitem[Rezaeian et~al.(2025)Rezaeian, Heidari, Azad, Merhof, Soltanian-Zadeh,
  and Hacihaliloglu]{rezaeian2025sl2a}
Rezaeian, R.; Heidari, M.; Azad, R.; Merhof, D.; Soltanian-Zadeh, H.; Hacihaliloglu, I. SL2A-INR: Single-Layer Learnable Activation for Implicit Neural Representation. In Proceedings of the IEEE/CVF International Conference on Computer Vision (ICCV),  Honolulu, HI, USA, 19--23 October 2025; pp. 26065--26074.

\bibitem[Vinogradskiy et~al.(2013)Vinogradskiy, Castillo, Castillo, Tucker, and
  Guerrero]{vinogradskiy2013ventilation}
Vinogradskiy, Y.; Castillo, R.; Castillo, E.; Tucker, S.; Guerrero, T. Use of 4-Dimensional Computed Tomography-Based Ventilation Imaging to Correlate Lung Dose and Function With Clinical Outcomes. \emph{Int. J. Radiat. Oncol. Biol. Phys.} \textbf{2013}, \emph{86}, 366--371. \url{https://doi.org/10.1016/j.ijrobp.2013.01.004}.

\bibitem[Qiao et~al.(2020)Qiao, Wang, Guo, Huang, Xia, and
  Tao]{Qiao2020registration}
Qiao, M.; Wang, Y.; Guo, Y.; Huang, L.; Xia, L.; Tao, Q. Temporally coherent cardiac motion tracking from cine MRI: Traditional registration method and modern CNN method. \emph{Med. Phys.} \textbf{2020}, \emph{47}, 4189--4198. \url{https://doi.org/10.1002/mp.14341}.

\bibitem[Reuter et~al.(2012)Reuter, Schmansky, Rosas, and
  Fischl]{reuter2012within}
Reuter, M.; Schmansky, N.J.; Rosas, H.D.; Fischl, B. Within-subject template estimation for unbiased longitudinal image analysis. \emph{NeuroImage} \textbf{2012}, \emph{61}, 1402--1418. \url{https://doi.org/10.1016/j.neuroimage.2012.02.084}.

\bibitem[Ulyanov et~al.(2018)Ulyanov, Vedaldi, and Lempitsky]{ulyanov2018deep}
Ulyanov, D.; Vedaldi, A.; Lempitsky, V. Deep Image Prior. In Proceedings of the IEEE/CVF Conference on Computer Vision and Pattern Recognition (CVPR), Salt Lake City, UT, USA, 18--23 June 2018; pp. 9446--9454.

\end{thebibliography}

\PublishersNote{}
\end{adjustwidth}
\end{document}